\documentclass{TST}

\TSTsetup{
type      = {Research Article},
doi       = {xxxx},
title     = {Practical Quantum CIM Empowerment \textit{via} All-Domestic-Core Agentic Large Model},
author    = {Wang Rui\cor{} and Lu Diannan\cor{}},
abstract  = {
Quantum computing devices are recognized as powerful tools for solving NP-complete problems. However, the intricacy of their modeling presents notable barriers for non-specialists, while the tedious iteration of constraint weights and modeling methodologies also consumes substantial effort on the part of experts. To address these challenges, this study integrates a femtosecond laser-pumped Coherent Ising Machine (CIM) with an LLM-driven agentic system by leveraging the LangGraph and LangChain frameworks. Comprehensive investigations demonstrate that large language models (LLMs) can effectively perform such tasks in modeling as QUBO/Ising model calibration, constraint weight decision iteration and rapid validation of literature-reported schemes. Notably, all these tasks can be fully implemented based on domestic large models, combined with domestically developed CIM hardware, we truly achieve the practical empowerment of quantum CIM that fully relies on all-domestic agentic large models and hardware. This work successfully realizes robust technological integration, laying a solid foundation for subsequent research. Nevertheless, it also identifies the persisting challenges in the two cutting-edge fields of large models and quantum computing at the current stage. Encouragingly, we unexpectedly discover a promising new paradigm where accumulated knowledge from agent‑assisted quantum computing iterations reciprocally enhances the agent's own problem-solving capability, thereby addressing these challenges.
},
keywords  = {Quantum Computing, Coherent Ising Machine (CIM), Agentic Large Models (ALMs), QUBO/Ising Model, Domestic Core},
received  = 20xx-xx-xx,
revised   = 20xx-11-xx,
accepted  = 20xx-00-xx,
}

 \address{Wang Rui and Lu Diannan are with the Department of Chemical Engineering, Tsinghua University, Beijing, 100084, China. \\
 E-mail: r-wang18@tinghua.org.cn;  ludiannan@mail.tsinghua.edu.cn}

\begin{document}

\maketitle

\section{Introduction}

Quantum computing stands as one of the most disruptive computational paradigms in the post-Moore era \cite{1, 2}, demonstrating potential far beyond classical computers in addressing specific computational challenges such as combinatorial optimization, cryptanalysis and materials simulation \cite{3,4,5,6,7,8,9,10,11,12,13,14,15,16,17,18}. Particularly in solving NP-complete problems, quantum annealing and specialized quantum simulators based on the Ising model have become focal points of pursuit in both academia and industry  \cite{19,20,21,22,23}. Among these, the Coherent Ising Machine (CIM) pumped by femtosecond lasers, leveraging the unique physical mechanism of optical pulse degenerate oscillation cavities, can efficiently map and solve for the ground state energy of large-scale combinatorial optimization problems \cite{24,25,26,27}. Consequently, it is widely regarded as one of the most promising candidate tools for overcoming the bottlenecks of classical computing.

Nevertheless, the transition of quantum CIMs from physical hardware to widespread industrial application continues to encounter significant non-technical barriers and practical challenges. First, a prerequisite for solving problems with a CIM is the translation of real-world constraints—such as logistics routing, financial asset allocation, or drug molecule docking energies—into an abstract Quadratic Unconstrained Binary Optimization (QUBO) model or an Ising model \cite{28,29,30,31,32,33,34,35}. This modeling process relies heavily on profound domain knowledge and quantum physical intuition, posing an exceptionally high barrier to entry for industry experts lacking a background in quantum computing. Second, even for experienced quantum computing specialists, the meticulous calibration and iterative tuning of constraint weights tailored to specific CIM hardware characteristics consume substantial trial-and-error time and computational resources. A tangible manifestation of this inefficiency is the latency incurred during the invocation of actual quantum hardware for solving \cite{36}, coupled with the potential squandering of quantum resources through parallel trial-and-error cycles (Fig. 1A).
\begin{figure*}[t!]
\centering
\includegraphics[width=0.9\textwidth]{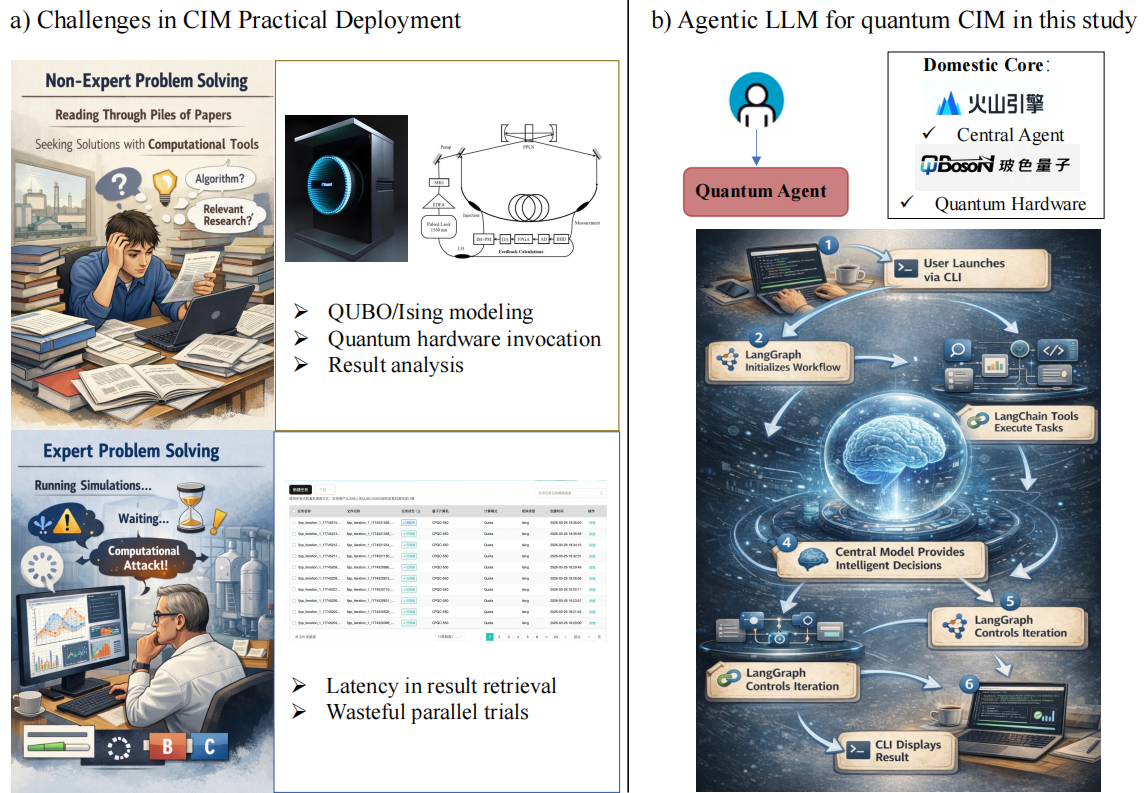}  
\caption{Quantum CIM Empowerment \textit{via} Agentic Large Model}
\label{fig1}
\end{figure*}

In recent years, agentic systems driven by Large Language Models (LLMs) have offered a novel avenue for addressing the aforementioned challenges of automated modeling and parameter iteration, owing to their robust capabilities in semantic understanding, logical reasoning, and code generation \cite{37,38,39,40,41,42,43}. While international research has explored the use of LLMs to assist in quantum circuit design \cite{44,45,46,47}, the domain of automated empowerment for specialized quantum simulators—particularly Coherent Ising Machines—remains largely uncharted. More critically, against the backdrop of intensifying global technological competition, establishing a fully self-reliant ‘‘Large Model + Quantum" technological closed-loop, grounded entirely in domestic computing infrastructure and independently controllable core hardware, holds inestimable value for safeguarding national security and strategic leadership in next-generation technology.

To address these challenges and realize this application vision, we previously validated in our workflow that the KaiwuAPI, which facilitates access to the physical CIM quantum hardware of \textit{QBoson Quantum Technology Co.}, can be seamlessly integrated into Markov decision workflows \cite{48}. Building upon this foundation, this study proposes and implements, for the first time, an empowerment framework for practical quantum CIM driven by an all-domestic-core agentic large model system (Fig. 1B). Specifically, the principal innovative contributions of this work are as follows: First, we employ the LangGraph and LangChain frameworks  \cite{49} to construct an agentic workflow endowed with autonomous planning and self-reflection capabilities that supports interactive operation \textit{via} a Command-Line Interface (CLI), thereby enabling cognitive automated generation spanning from natural language requirement descriptions to QUBO/Ising model calibration. Second, we have successfully achieved deep adaptation and integration between the aforementioned domestic large model agentic system and domestically developed femtosecond laser-pumped CIM hardware. Through this software-hardware co-design, we not only substantially lower the barrier for non-specialists to access quantum computational power but also significantly reduce the modeling and tuning cycles for expert users. Third, we validate the proficiency of domestic large models in comprehending complex physical constraints, executing iterative constraint weight decisions, and reproducing and validating classical schemes reported in the literature, thereby substantiating the usability and reliability of domestic foundation models in quantum computing auxiliary tasks.

Finally, based on the unexpected turnaround observed in practical quantum computing tasks executed by the agent, we decoded a novel quantum-enhanced agent paradigm that transcends the pre-set quantum-agent integration mode. The quantum computing iteration process provides rich task feedback and high-quality contextual guidance for the central large language model. With such iterative experience, the model can autonomously optimize suboptimal solutions to the global optimum. This discovery establishes a new agent development path of "quantum iteration empowering large language model reasoning", which features stronger robustness and breaks through the limitations of traditional hardware-only acceleration and strict precise solving requirements.

\section{Methods}

\begin{figure*}[t!]  
\centering
\includegraphics[width=0.75\textwidth]{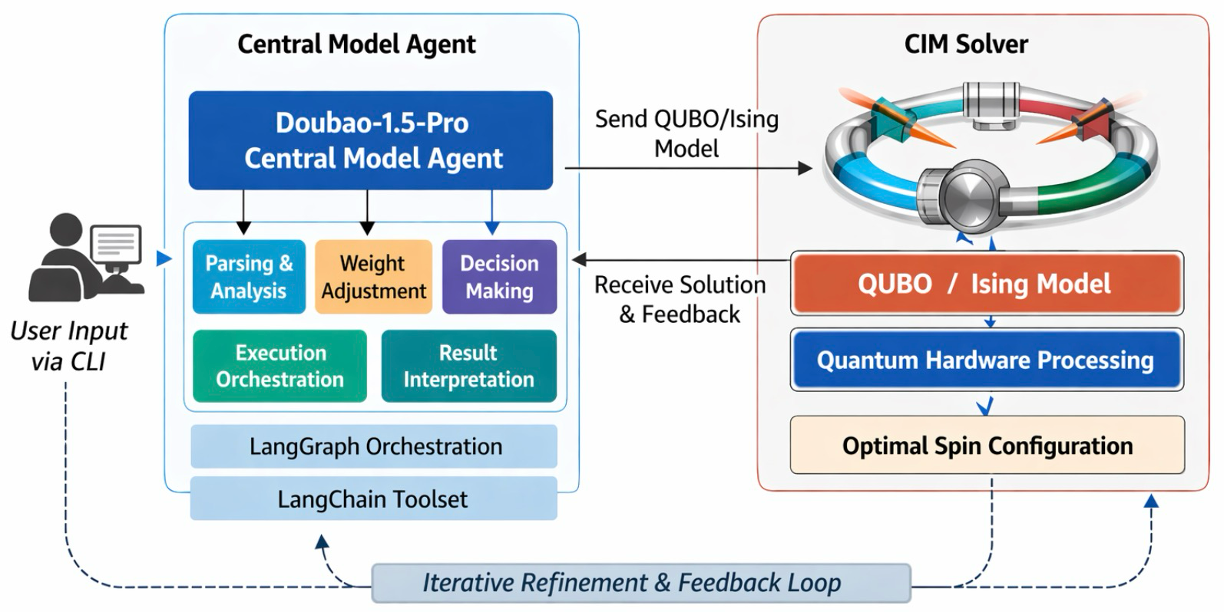}  
\caption{General System Architecture of the CIM Empowerment Framework}
\label{fig2}
\end{figure*}
The quantum Coherent Ising Machine (CIM) empowerment framework proposed in this study is constructed upon a comprehensive comparative evaluation of multiple domestic and international large language models. At its core, the system employs the domestic \textit{Doubao-1.5-Pro} central large model agent as the primary decision-making unit. This agent is deeply coupled and synergistically operated with several peripheral modules—namely, a Command-Line Interface (CLI) interaction portal, LangGraph workflow orchestration, and LangChain tool encapsulation—to constitute a fully functional quantum optimization agentic system. All peripheral modules are architected specifically around the decision-making requirements of the central agent, thereby realizing a closed-loop integration of command transmission, process governance, tool invocation and intelligent reasoning (Fig. 2). The detailed implementation and cooperative mechanisms are delineated as follows.

\subsection{Core Unit: The Central Model Agent}
The Central Model Agent serves as the cognitive and decision-making hub of the entire system, implemented within the module  \texttt{nodes/center\_model\_v5.py}. This module encapsulates the Volcano Engine API at its foundation and relies on the central large model to execute global intelligent decision-making. The foundational implementation of this agent extends beyond mere reception of holistic process state information; it also conducts analytical feedback based on the standardized textual results returned from quantum hardware execution. Consequently, it governs the core logic of quantum optimization: performing problem context parsing grounded in natural language requirements and associated environmental documentation; executing constraint satisfaction analysis in conjunction with CIM solution outputs; generating targeted adjustment schemes for QUBO/Ising model constraint weights; and outputting decision rationales accompanied by confidence assessments to provide pivotal directives for iterative refinement. Critical capabilities of the system—including adaptive optimization and fine-grained model calibration—are directly driven by this Central Agent, establishing it as the essential vector for realizing ‘‘Large Model + Quantum" intelligent empowerment.
\subsection{Interaction Portal: CLI}
The CLI module (\texttt{cli.py}) functions as the interactive entry point between the system and the user, dedicated entirely to facilitating task initiation and result feedback for the Central Model Agent. Its primary role is to receive external directives on behalf of the central agent: it parses user-input parameters such as natural language problem descriptions and maximum iteration thresholds, transmitting standardized task instructions to the central agent. Concurrently, it receives decision-making progress updates, iteration states, and final solution outcomes from the central agent, rendering them visually to the user. Furthermore, the CLI supports batch processing and automated execution, providing a convenient interactive conduit for the central agent to perform quantum optimization tasks serially or in parallel. It constitutes the direct bridge connecting the human operator to the core decision-making intelligence.
\subsection{Workflow Orchestration: LangGraph}
The LangGraph module, instantiated \textit{via} the QuantumAgent class within \texttt{agent.py}, serves as the execution governance skeleton and \textbf{ReAct reasoning core} of the Central Model Agent, operationalizing the agent's decision workflow through the standard reasoning-acting (ReAct) paradigm \cite{50}. This module defines eight core execution nodes centered around the central agent's logic: Parsing, Modeling, Solving, Result Interpretation, Evaluation, Memory, Decision-Making and Weight Adjustment. These nodes are strictly partitioned into two categories in accordance with the ReAct framework: reasoning nodes (Parsing, Evaluation, Memory, Decision-Making) for logical analysis and judgment, and acting nodes (Modeling, Solving, Result Interpretation, Weight Adjustment) for practical execution. The execution logic of all nodes is strictly oriented toward the directives issued by the central agent. Specifically, the module controls the execution sequence and inter-node dependencies based on the agent's planning; maintains global state variables—including iteration history and incumbent optimal solutions—and synchronizes them with the central agent; regulates the iterative cycle according to termination conditions adjudicated by the central agent; and provides real-time feedback of the workflow execution trajectory to the central agent, thereby furnishing contextual evidence for subsequent decisions. This ensures orderly governance and robust state support for the decision-making processes of the central agent.

\subsection{Tool Execution Backend: LangChain}
The LangChain tool module (\texttt{tools/quantum\_tools.py}) constitutes the \textbf{ReAct acting execution layer} and standardized execution interface, through which the Central Model Agent invokes quantum computational capabilities. It is specifically designed to underpin the quantum-related decision-making of the central agent by translating high-level reasoning instructions into low-level executable operations. This module encapsulates core quantum operations—including QUBO/Ising model construction, invocation of physical CIM quantum hardware, and result parsing—as standardized tool abstractions. It receives execution commands dispatched by the central agent and performs the concrete quantum computational tasks. During execution, it captures anomalies and relays execution outcomes back to the central agent, supplying data substantiation for the next round of decision-making. Additionally, the module supports toolchain reusability and combinatorial assembly, enabling flexible adaptation to the diverse decision-making requirements of the central agent. This architecture facilitates seamless interfacing between low-level quantum capabilities and high-level large model reasoning, and completes the closed-loop of the ReAct paradigm.

\subsection{CIM Hardware and QUBO/Ising Modeling}
The physical computing substrate of the proposed framework is the Coherent Ising Machine (CIM) developed by  \textit{QBoson Quantum Technology Co.} Unlike gate-based quantum computers, the CIM is a specialized quantum simulator that exploits the spontaneous symmetry breaking of degenerate optical parametric oscillators (DOPOs) pumped by femtosecond laser pulses within a fiber ring cavity. In this architecture, each DOPO pulse corresponds to an Ising spin variable \( s_i \in \{-1, +1\} \), and the mutual coupling among pulses is implemented  \textit{via} optical delay lines and phase-sensitive amplification. The system naturally evolves toward the configuration that minimizes the overall optical loss, which corresponds precisely to the ground state of the encoded Ising Hamiltonian.

To map a constrained combinatorial optimization problem onto the CIM hardware, the problem must first be cast into the Quadratic Unconstrained Binary Optimization (QUBO) formalism. A generic QUBO problem is defined as the minimization of a quadratic form over binary variables \( \mathbf{x} \in \{0, 1\}^n \):
\begin{equation}
\min_{\mathbf{x} \in \{0,1\}^n} \sum_{i=1}^n Q_{ii} x_i + \sum_{i<j} Q_{ij} x_i x_j = \mathbf{x}^T Q \mathbf{x}
\label{eq:qubo_full_form}
\end{equation}

where \( Q \in \mathbb{R}^{n \times n} \) is an upper-triangular matrix encoding both linear coefficients (diagonal elements \( Q_{ii} \)) and quadratic interaction terms (off-diagonal elements \( Q_{ij} \)).
\subsubsection{Incorporation of Constraints and Penalty Weights}

Since the CIM natively solves \textit{unconstrained} optimization problems, logical constraints must be reformulated as penalty terms appended to the objective function. Consider a problem with an original objective function \( f(\mathbf{x}) \) and a set of constraints \( \{C_k(\mathbf{x}) = 0\}_{k=1}^m \). For each constraint \( C_k(\mathbf{x}) \), a quadratic penalty term weighted by a scalar coefficient \( \lambda_k > 0 \) is added to the objective, yielding the penalized QUBO cost function:
\begin{equation}
H_{\mathrm{QUBO}}(\mathbf{x}) = f(\mathbf{x}) + \sum_{k=1}^m \lambda_k \cdot \left(C_k(\mathbf{x})\right)^2
\label{eq:penalized_qubo}
\end{equation}

The parameters \( \lambda_k \) are precisely the \textbf{constraint weights} that govern the trade-off between objective minimization and constraint satisfaction. In matrix form, the penalty terms contribute additional entries to the \( Q \) matrix. Specifically, if the constraint \( C_k(\mathbf{x}) \) expands to a linear or quadratic expression in \( \mathbf{x} \), its square yields quadratic and linear terms that are absorbed into \( Q \) with coefficients proportional to \( \lambda_k \).

The QUBO problem is isomorphic to the Ising model  \textit{via} the affine transformation \( s_i = 2x_i - 1 \), mapping \( x_i \in \{0, 1\} \) to spin variables \( s_i \in \{-1, +1\} \). Substituting \( x_i = (s_i + 1)/2 \) into the QUBO cost function yields the Ising Hamiltonian:
\begin{equation}
E(\mathbf{s}) = \sum_{i=1}^n h_i s_i + \sum_{i < j} J_{ij} s_i s_j + \text{constant}
\label{eq:ising_hamiltonian_qubo}
\end{equation}

The local fields \( h_i \) and coupling strengths \( J_{ij} \) are linear combinations of the QUBO matrix entries, and therefore inherit the constraint weights \( \lambda_k \). Specifically, the penalty terms contribute to the Ising parameters as follows:

\begin{equation}
h_i = -\frac{1}{2} \left( Q_{ii} + \sum_{j \neq i} Q_{ij} \right), \quad J_{ij} = \frac{1}{4} Q_{ij} \quad (i < j)
\label{eq:qubo_to_ising_mapping}
\end{equation}

\begin{figure*}[t!]  
\centering
\includegraphics[width=0.85\textwidth]{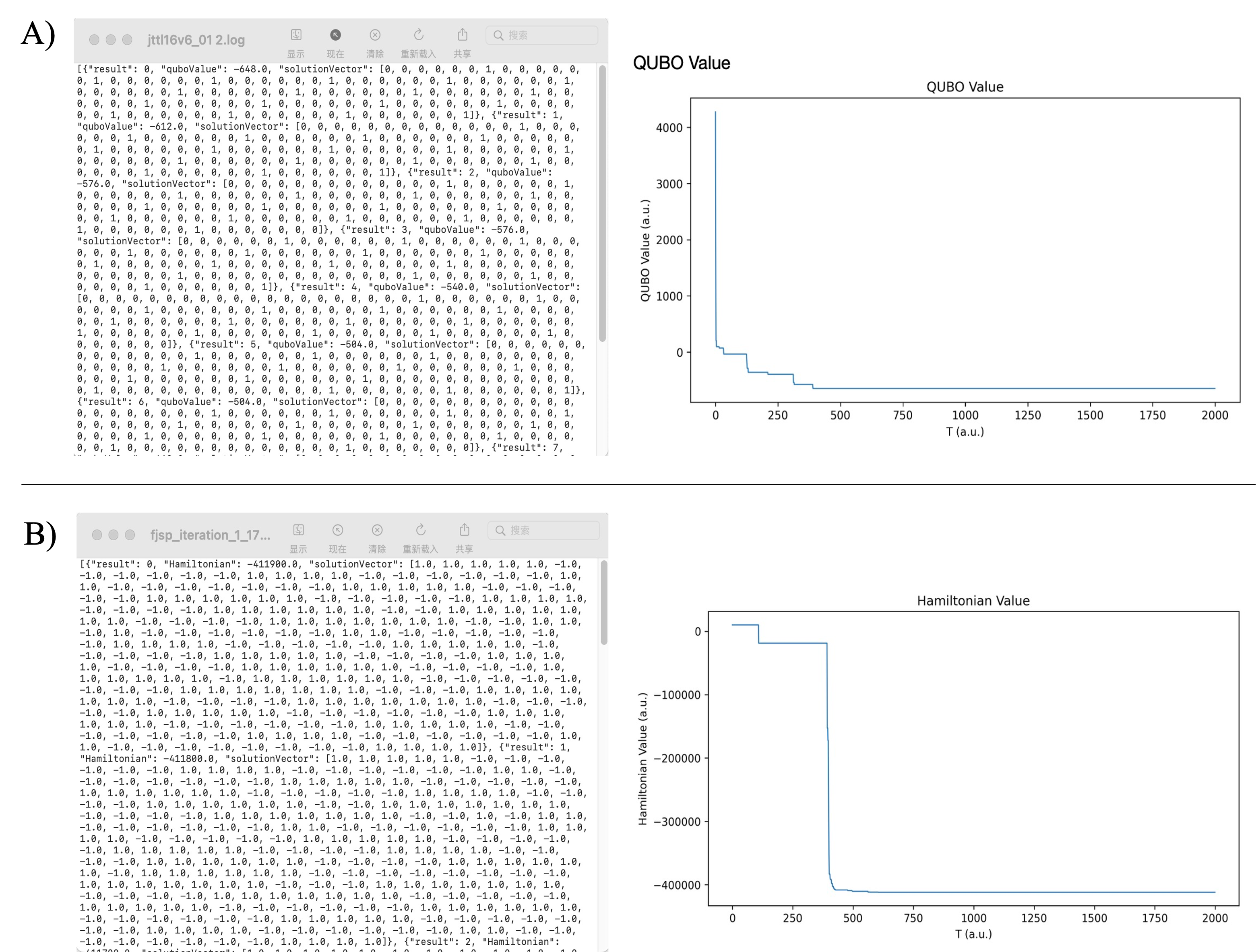}  
\caption{QUBO(A) and Ising(B)  Solution Outputs from CIM }
\label{fig3}
\end{figure*}

Consequently, adjusting a constraint weight \( \lambda_k \) modifies both the local fields and the pairwise couplings sent to the CIM hardware. The magnitude of \( \lambda_k \) determines the relative strength of the penalty-induced interactions compared to the original objective terms. An improperly chosen \( \lambda_k \) can either fail to enforce the constraint (if too small) or dominate the energy landscape, leading to poor objective values and potential precision saturation of the CIM's analog coupling parameters (if too large). The QBoson CIM platform exposes the KaiwuAPI, which accepts either the QUBO matrix \( Q \) or the Ising parameters \( (h, J) \) and returns the measured spin configuration along with the corresponding energy.

\subsection{Integrated Execution Workflow}
The entire system operates with the Central Model Agent as the cognitive driver, where the physical QBoson CIM hardware and the iterative refinement of constraint weights form a tightly coupled feedback loop. The fusion of hardware and agentic intelligence manifests throughout the execution pipeline.

\subsubsection{Feasibility of the Integrated Workflow}

As verified in our previous project work, the invocation of the physical CIM hardware via KaiwuSDK and KaiwuAPI features highly standardized and fixed formats: the input strictly follows the unified specification of QUBO matrix or Ising parameters, and the output is limited to no more than ten optimal solution vectors with a consistent structural format (Fig.3). This high standardization of both input and output enables the direct embedding of hardware invocation and result parsing modules into the Markov decision workflow, simplifies agent-side result assessment and iteration, and provides a robust technical basis for realizing the integrated execution pipeline.

\subsubsection{Necessity of Core Model}

Preliminary evaluation of the central model indicates that open-source small models, such as Qwen2.5-14B, perform adequately on classical combinatorial optimization tasks (e.g., TSP, max-cut) under simplified settings. However, real-world applications entail substantially greater complexity—including sophisticated modeling, precision tuning of matrix parameters for physical CIM hardware, in-depth solution analysis, and closed-loop iterative refinement. These demands necessitate the superior reasoning capabilities, domain adaptability, and generalization afforded by closed-source large language models. Moreover, the central model's intelligent document parsing and multi-turn dialogue reasoning mechanisms are instrumental in enabling rapid literature-driven modeling and validation workflows. Concurrently, agentic assistance proves essential in balancing the inherent trade-off between runtime latency and the inefficient consumption of parallel computational resources during iterative processes. 

Among the models evaluated, Gemini-3-Pro and Doubao-1.5-Pro demonstrate the most favorable and consistently stable performance. Although domestic large models still exhibit certain relative shortcomings—most notably in prompt-cache utilization—they nonetheless offer a distinct token-cost advantage in the context of the tasks examined in this study. Without recourse to distilling large-model capabilities into smaller open-source alternatives, and with the objective of expeditiously establishing a fully self-reliant domestic infrastructure, this work adopts Doubao-1.5-Pro as the core reasoning engine, thereby achieving full-domestic-core empowerment for quantum CIM hardware.

\subsubsection{Multi-turn Dialogue Reasoning}
 
 This work incorporates a mechanism of intelligent document parsing and multi-turn dialogue reasoning driven by the central model agent and PromptTemplate tools within LangChain into the overall framework:
\begin{enumerate}
\item Multi-turn dialogue reasoning belongs to the internal cognitive reasoning layer of the central model, distinct from the outer CLI interaction layer. Through continuous API calls to the central large model, it accomplishes in-depth document analysis, code-aware context understanding, iterative optimization of modeling logic, and stepwise rule verification, with the central model agent and the LangChain toolset as the interactive subjects.
\item Execution pipeline: In the first round, basic document information and analysis objectives are input to the central model; in the second round, the model outputs preliminary analysis results and identifies contents requiring further investigation; in the third round, detailed inquiries are conducted on key issues such as weight parameter selection; in subsequent rounds, in-depth analysis continues based on model feedback until complete modeling information is obtained.
\item Hierarchy and data flow: The CLI serves as the outer interaction portal for user–system communication; multi-turn dialogue acts as the inner reasoning core to support the central model in document analysis. In addition to natural language documents, the input context may also include structured descriptions or excerpts of the underlying hard-coded modeling implementations. This enables the central model to perform code-aware reasoning, identify adjustable parameters, and align its decisions with the predefined computational workflow. The overall non-cyclic data flow (without ReAct) is User → CLI → Agentic Workflow → Multi-turn Dialogue Reasoning of the Central Model → Intelligent Document Parsing → Iterative Constraint Weight Tuning → QUBO/Ising Modeling → CIM Hardware Solving → Result Output → CLI Feedback to User.

\end{enumerate}

\subsubsection{Parameter Passing Mechanism}

\begin{figure}[h!t]
\centering
\includegraphics[width=1.0\linewidth]{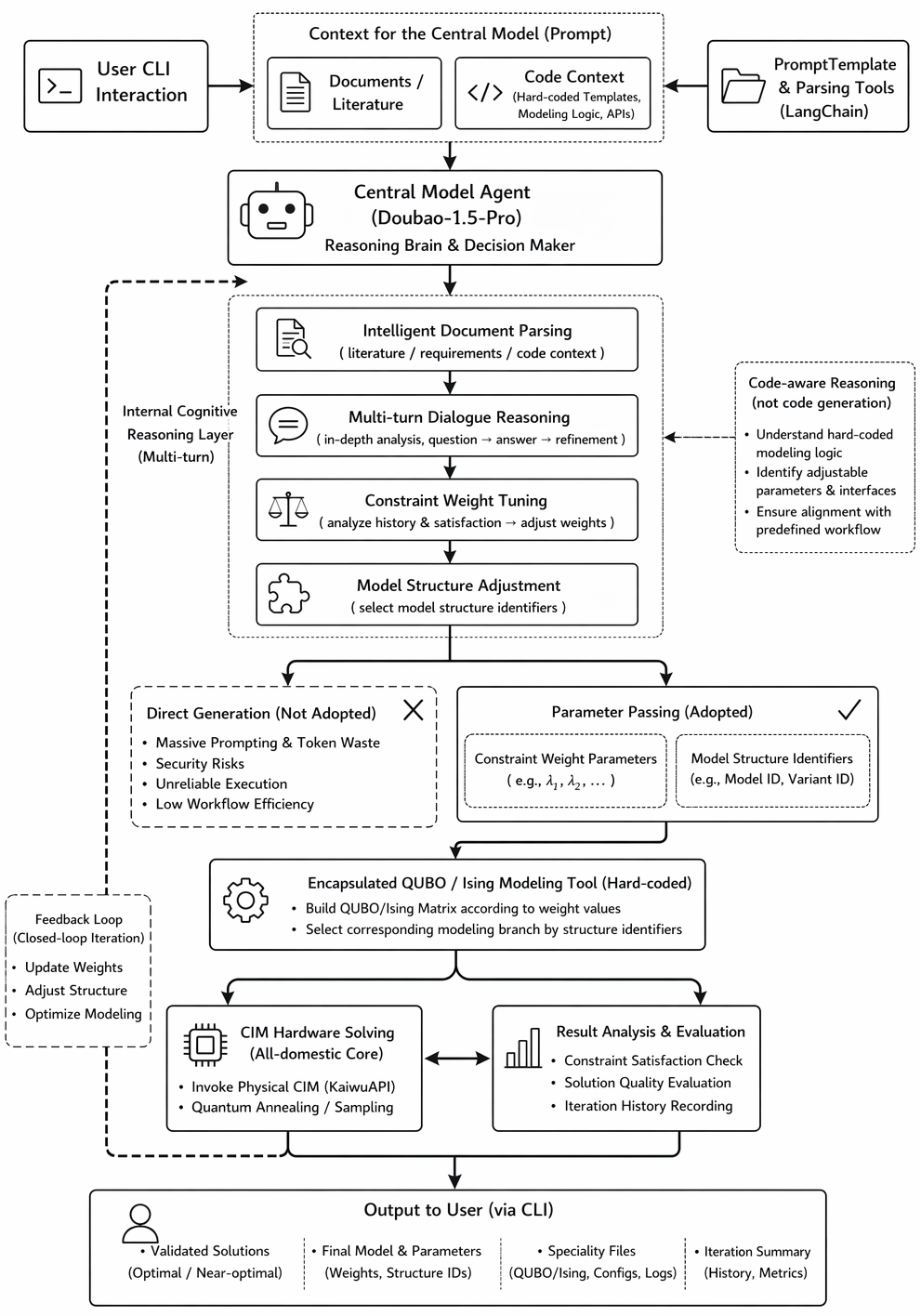}  
\caption{Agentic Workflow for QUBO/Ising Modeling.}
\label{fig4}
\end{figure}

Within the LangChain framework including Multi-turn Dialogue Reasoning, we also tested the central model’s ability to directly construct QUBOs from scratch or modify and upload solver input files for the physical CIM using tools such as PythonREPLTool. Unfortunately, all closed-source models showed clear limitations in this process. To ensure these models correctly perform such complex operations, extensive and detailed prompt engineering was required, which resulted in massive and inefficient token consumption, greatly raising inference costs and reducing workflow efficiency. After comprehensively evaluating the practical necessity, associated security risks, and severe token waste caused by excessive prompt guidance, we decided to hard-code most of the nodes for QUBO modeling. The central model interacts with these predefined modules in a code-aware manner, and synchronously performs only two core tasks \textit{via} a unified parameter passing mechanism: constraint weight tuning and model structure adjustment (see Fig. 4 for the full workflow, the low-level matrix construction is handled by an encapsulated tool \textit{via}  parameter-passing).

In the constraint weight tuning step, the central model analyzes the iteration history, CIM solving results, and constraint satisfaction status, and then outputs adjustment suggestions for the constraint weight parameters. In the model structure adjustment step, the central model outputs model structure identifier parameters. The above parameters are passed to the encapsulated QUBO/Ising modeling tool \textit{via} the LangChain tool invocation interface. The modeling tool constructs the QUBO/Ising matrix based on the weight values and selects the corresponding modeling logic branch according to the model structure identifiers.

\subsubsection{Full Execution Pipeline}
The full execution pipeline of the system is detailed as follows:
\begin{enumerate}
\item The user submits a natural language problem description \textit{via} the CLI to the Central Model Agent.
\item The central model conducts in-depth document analysis and extracts key information required for modeling \textit{via} the multi-turn dialogue reasoning mechanism.
\item LangGraph constructs the execution graph, wherein the \textit{Modeling} and \textit{Solving} nodes are explicitly linked to the hardware execution phase.
\item The Central Agent dispatches a modeling directive. The LangChain toolset constructs the QUBO matrix (or converts it to Ising parameters) and invokes the QBoson CIM through the KaiwuAPI. The physical optical pulses within the CIM cavity undergo annealing and settle into a ground-state configuration.
\item The Central Model Agent performs intelligent analysis based on tool execution feedback, outputting decisions pertaining to weight adjustment and model refinement.
\item LangGraph manages the iterative process based on the central agent's commands until predefined termination criteria are satisfied.
\item The CLI receives the final output from the central agent and presents the synthesized results to the user.
\end{enumerate}

The core advantage of this integrated architecture lies in its realization of holistic synergy with the central agent acting as the pivot. The modular design ensures that individual components can be maintained and upgraded independently while all peripheral modules revolve around the core decision unit. Strong observability enables the central agent to maintain real-time situational awareness of the entire process state. Standardized interfaces facilitate efficient interoperability between the decision layer and the execution layer. Ultimately, under the stewardship of the Central Model Agent, the system accomplishes full-process intelligent empowerment, spanning from natural language requirement specification to iterative quantum solution optimization.

\section{Results}

\subsection{Agent-Based QUBO Solution for Combinatorial Task}

To validate the problem-solving capability of the proposed agentic framework, we design a mass spectrometry-inspired combinatorial optimization task. Given a measured peptide/protein mass, the agent infers its amino acid composition through QUBO modeling. Mathematically, this problem is the \textbf{Unbounded Subset Sum Problem} and \textbf{Approximate Integer Solution to Linear Diophantine Equations}, formulated as:
\[
\sum_{i} n_i m_i \approx M_{\text{target}}
\]

where $n_i \in \{0,1,2,\dots\}$ denotes amino acid counts, $m_i$ is the monoisotopic mass of the $i$-th amino acid, and $M_{\text{target}}$ is the measured mass. Mass calibration is performed as $M_{\text{cal}} = M_{\text{target}} - M_{H_2O}$, and each amino acid mass is adjusted by half a water molecule for peptide bond formation. 

Through hard-coding implementation, agent preparation, and partial auxiliary iterative verification (detailed in the Appendix), we adopt the position-based one-hot encoding modeling approach. We defined the binary variable $x_{s,a} \in \{0,1\}$ which means that the $s$-th position of the sequence selects the $a$-th amino acid, reconstructed the QUBO model with the position one-hot constraint and the total mass matching constraint the position one-hot constraint ensured that each position has and only one amino acid selected, and its penalty term is:
\[
\lambda_{\text{pos}} \sum_{s=1}^{S} \left(1 - \sum_{a} x_{s,a}\right)^2
\]

where $\lambda_{\text{pos}}$ was the penalty weight of the position constraint, the total mass matching constraint minimized the deviation between the total mass of the sequence and the target mass, and its penalty term was:
\[
\lambda_{\text{mass}} \left(\sum_{s=1}^{S} \sum_{a} m_a x_{s,a} - M_{\text{target}}\right)^2
\]

where $\lambda_{\text{mass}}$ was the penalty weight of the mass constraint, the total QUBO model was the sum of the two penalty terms, namely:
{\footnotesize
\[
H = \lambda_{\text{pos}} \sum_{s=1}^{S} \left(1 - \sum_{a} x_{s,a}\right)^2 + \lambda_{\text{mass}} \left(\sum_{s=1}^{S} \sum_{a} m_a x_{s,a} - M_{\text{target}}\right)^2
\]
}

This modeling approach decouples count-based encoding from mass parameters, avoids scaling-induced coefficient suppression, and preserves clear physical interpretability; the agent automates optimized QUBO matrix construction (via weight adjustment and hard-coded modeling selection), quantum hardware invocation, full CIM solution analysis, and closed-loop constraint weight tuning, ultimately yielding stable and reasonable solutions. Building on this model, the position-based one-hot QUBO framework resolves the coefficient suppression issue caused by the coupling of amino acid mass and encoding weights in early count-based encodings, with its explicit matrix structure supporting direct preprocessing and input into physical CIM hardware for solution. The central large model agent autonomously conducts closed-loop iterative optimization of constraint weights, performs systematic cognitive solution validation on the physical CIM, and identifies core hardware-level performance bottlenecks from measured results—completing solution parsing, constraint satisfaction evaluation, matrix coefficient diagnosis, and weight parameter decision-making independently, without manual intervention, as the underlying code only receives standardized parameters from the agent to update the model and invoke the hardware. The core objective of weight iteration is to balance position one-hot constraint satisfaction and total mass matching accuracy within the one-hot encoding framework, while adapting the QUBO matrix to the input range of the physical CIM hardware, and the agent has designed and implemented a gradient-based iterative workflow to achieve this goal.

\begin{enumerate}
    \item \textbf{Baseline Test with Equal Weight Initialization}
    The agent performed an initial baseline validation with equal weights $\lambda_{\text{pos}} = \lambda_{\text{mass}} $ for the target peptide LACRP4 (real sequence: KKSKAKEPPPKKT; target mass: 1448.77 Da after dehydration \cite{51} ). Among the top 10 optimal solutions returned by the physical CIM, 9 exhibited one-hot constraint violations, corresponding to a violation rate of 90\%: 7 solutions selected multiple amino acids at a single position, and 2 solutions selected no valid amino acids at any position. Meanwhile, the total mass of the optimal solution deviated from the target by 312.4 Da, failing to meet basic task requirements.

    \item \textbf{Autonomous Gradient Tuning of Weight Ratios}
    To address the constraint imbalance observed in the baseline test, the agent designed an 8-round gradient-based tuning experiment covering $\lambda_{\text{pos}}:\lambda_{\text{mass}}$ from $1:10$ to $8.5\times10^4:1$. Two typical failure modes were identified:
    \begin{itemize}
        \item When $\lambda_{\text{mass}} \geq \lambda_{\text{pos}}$ (mass-dominant, ratio = $1:10$), the mass-matching term dominated the energy function, the one-hot constraint penalty contributed less than 1\%, and the violation rate remained above 85\%, resulting in invalid sequences;
        \item When $\lambda_{\text{pos}} \gg \lambda_{\text{mass}}$ (position-dominant, ratio $> 10^4:1$), the violation rate dropped below 12\%, but mass-matching accuracy deteriorated sharply, with a maximum deviation of 427.3 Da. The solver tended to select amino acids with the smallest molecular weights (Gly, Ala) to minimize the energy value, losing mass-matching capability.
    \end{itemize}

    \item \textbf{Supplementary Validation of Encoding Stability}
    
    The agent further conducted parallel validation of three alternative encoding schemes: 3/2-bit binary count encoding, equal-weight count encoding, and mass-unit integer encoding. Results showed that all count-based encodings suffered from severe coefficient suppression: after normalization, linear and quadratic terms were compressed by up to 42.5 times, with most coefficients approaching zero, and the one-hot constraint violation rate generally exceeded 92\%. This further validates that position-based one-hot encoding provides the most stable constraint expression and clear physical interpretability.
\end{enumerate}

After iterative optimization in numerical simulation, the agent confirmed through multiple rounds of physical CIM measurements that accurate and stable solving remains unachievable due to the following key bottlenecks:

\begin{enumerate}
    \item \textbf{Numerical Precision Loss from 8-bit Integer Quantization}
    
    The physical CIM hardware requires the input QUBO matrix to be converted into 8-bit signed integers (range: $-128 \sim 127$), which involves linear scaling and integer rounding. The original floating-point QUBO matrix spans six orders of magnitude. After linear scaling, small coefficients are rounded to zero, losing all valid constraint information. Quantization-induced rounding errors severely distort the relative weights of constraints and break the well-tuned weight balance from simulation, resulting in constraint failure during hardware solving.

    \item \textbf{Inherent Noise of Physical CIM Hardware}
    
    The femtosecond laser-pumped physical CIM introduces inherent quantum noise and analog circuit errors during optical pulse oscillation, phase-sensitive amplification, and spin-state readout, leading to random flipping of spin configurations in solutions (Although no significant numerical fluctuations were observed tduring additional multiple tests, it should still be fully considered when solving practical problems).

    \item \textbf{Loss of Precise Solving Capability}
    Limited by quantization precision loss and inherent hardware noise, the physical CIM cannot achieve accurate target mass matching. In all test rounds, the minimum mass deviation of valid solutions reached 52.3 Da (relative deviation $>3.6\%$), two orders of magnitude worse than the theoretical precision ($<0.1\%$) in numerical simulation. Moreover, the deviation in actual amino acid composition is far more severe than reflected by numerical values. The hardware can only output rough approximate solutions and cannot support precise solving under strict numerical constraints.
\end{enumerate}

Cognitive measurements by the agent validated that the current system cannot achieve accurate and stable solving of combinatorial optimization tasks under strict numerical constraints. The key limiting factors are severe precision loss caused by 8-bit integer quantization, inherent quantum noise, and analog errors of the physical hardware. Input quantization bit-width limitations and inherent physical noise represent the core bottlenecks restricting the practical deployment of physical CIM in fine-grained combinatorial optimization scenarios.

\subsection{Agent-Based Ising Solution for Autonomous Modeling and Literature Reimplementation}
In the previous case, autonomous QUBO model iteration by the agent preliminarily validated the feasibility of the agent’s automated participation in quantum computing decision-making iterations \textit{via} the ReAct workflow under the LangGraph framework. On this basis, we further realizes Ising model construction assisted by the agent for the reimplementation of literature-reported schemes. We select the FJSP (Flexible Job-shop Scheduling Problem) quantum solving framework proposed by Fu et al. \cite{24} as the benchmark task, a representative validation of quantum combinatorial optimization in industrial scheduling scenarios (Detailed FJSP Formulation, QUBO Modeling, Ising Conversion and Variable Pruning in this documents showed in Appendix). The core objective of this section is to verify the capability of the proposed agent system in autonomous literature parsing, error self-diagnosis, correction, and full-process parameter iteration.

\subsubsection{Benchmark FJSP Instance}
The FJSP instance is defined by a set of $A$ jobs $J = \{J_1, J_2, ..., J_A\}$ and a set of $B$ machines $M = \{M_1, M_2, ..., M_B\}$. Each job $j \in J$ consists of a sequence of $h_j$ operations, where the $h$-th operation of job $j$, denoted as $o_{j,h} \in O_j$, must be processed on one machine from the eligible machine set $M_{j,h} \subseteq M$. Each machine can process at most one operation at a time, and the processing time of an operation varies across different eligible machines. The core objective is to minimize the maximum completion time (makespan) of all jobs.

{\footnotesize
\begin{align*}
&J & &\text{Set of jobs} \\
&M & &\text{Set of machines} \\
&O_j & &\text{Set of operations in job } j \ (j \in J) \\
&M_{j,h} & &\text{Set of eligible machines for operation } o_{j,h} \\
&h_j & &\text{Number of operations in job } j \\
&o_{j,h} & &\text{The } h\text{-th operation of job } j \\
&p_{i,j,h} & &\text{Processing time of } o_{j,h} \text{ on machine } i \in M_{j,h} \\
&s_{j,h} & &\text{Start time of operation } o_{j,h} \\
&c_{j,h} & &\text{Completion time of operation } o_{j,h} \\
&T = \{0,1,...,T_{\text{max}}\} & &\text{Discretized timeline with maximum time } T_{\text{max}} \\
&L & &\text{A sufficiently large positive constant}
\end{align*}
}

In this work, we present a new canonical benchmark instance for the Flexible Job Shop Scheduling Problem (FJSP), which comprises 3 jobs and 3 machines. The processing time $p_{i,j,h}$ for each eligible machine are listed in Table 1.
\begin{table}[h!]
\centering
\caption{Processing time of the 3×3 FJSP instance}
\label{tab:appendix_3x3_fjsp}
\begin{tabular}{@{}cccccc@{}}
\toprule
Job & Operation & $M_1$ & $M_2$ & $M_3$ \\
\midrule
\multirow{3}{*}{Job 1} & $o_{1,1}$ & 3 & 4 & 5 \\
& $o_{1,2}$ & 4 & 3 & 6 \\
& $o_{1,3}$ & 5 & 2 & 4 \\
\midrule
\multirow{3}{*}{Job 2} & $o_{2,1}$ & 4 & 6 & 5 \\
& $o_{2,2}$ & 3 & 4 & 5 \\
& $o_{2,3}$ & 2 & 5 & 3 \\
\midrule
\multirow{3}{*}{Job 3} & $o_{3,1}$ & 2 & 4 & 3 \\
& $o_{3,2}$ & 5 & 3 & 4 \\
& $o_{3,3}$ & 3 & 6 & 2 \\
\bottomrule
\end{tabular}
\end{table}

\subsubsection{Autonomous Literature Parsing and Modeling Pipeline}
Leveraging the multi-turn dialogue reasoning and intelligent document parsing mechanism, the central model agent autonomously completes the full parsing of the benchmark literature, extracting the core QUBO model formulation, four penalty terms with their corresponding weight coefficients ($\alpha,\beta,\gamma,\delta$), the maximum time parameter $t_{\text{max}}$ (max-time), variable pruning rules, and the standard conversion specification from QUBO to the Ising model. Since the core modeling logic is implemented via hard-coding, the core purpose of its literature parsing is to provide a basis for code execution verification, parameter rationality analysis, and automatic problem diagnosis during human–computer interaction. The agent identifies all key externally adjustable parameters in the modeling and solving process based on the parsing results.

In this work, we utilize the ReAct workflow of the LangGraph framework to implement agent reasoning and process orchestration and employ the tool invocation interface of LangChain to achieve parameter interaction and tool scheduling. Meanwhile, we manually design standardized parameter passing mechanisms to adapt to diverse system interaction and deployment scenarios, as detailed below:
\begin{itemize}
    \item \textbf{Function Parameter Passing}: The central model outputs adjustment suggestions for $\alpha,\beta,\gamma,\delta$ and $t_{\text{max}}$, and the encapsulated Python modeling tool directly receives the values as function parameters to complete the QUBO matrix construction.
    \item \textbf{Configuration File Delivery}: The central model generates JSON/YAML configuration files containing weight and iteration parameters, and the underlying code updates the running settings via hot-loading the configuration.
    \item \textbf{Environment Variable Injection}: The central model sets environment variables to store key parameters, which are read and loaded by the Python code through \texttt{os.environ}.
    \item \textbf{API Interface Call}: The central model returns parameter suggestions \textit{via} REST API, and the modeling code invokes the API to obtain the latest configuration in real time.
    \item \textbf{Shared Memory Transfer}: The central model stores parameters in shared memory or message queues, and the modeling code reads and uses them directly from the memory space.
\end{itemize}

After comprehensive evaluation of implementation complexity, security, token consumption efficiency, and hardware invocation compatibility, the agent selects function parameter passing as the standard scheme for this task. All core nodes of QUBO/Ising modeling for the FJSP are implemented in hard-coded form. The agent only synchronously performs the following core tasks: literature and code parsing and alignment, solution vector analysis and diagnosis after CIM solving, constraint weight tuning, and model structure adjustment. All underlying operations are completed by encapsulated tools through standardized parameter input. All tasks in this section can be executed via the command-line interface (CLI) developed in this work, and typical invocation commands are as follows:

{
\footnotesize
\begin{verbatim}
# 3×3 FJSP Basic Usage
python cli.py --problem "3-job 3-machine FJSP" 
    -i * -t *

# Full Task Invocation Example
python cli.py -p "3-job 3-machine FJSP:
Job1: Op1:M1(3),M2(4),M3(5);Op2:M1(4),M2(3),M3(6);
       Op3:M1(5),M2(2),M3(4).
Job2: Op1:M1(4),M2(6),M3(5);Op2:M1(3),M2(4),M3(5);
       Op3:M1(2),M2(5),M3(3).
Job3: Op1:M1(2),M2(4),M3(3);Op2:M1(5),M2(3),M3(4);
       Op3:M1(3),M2(6),M3(2)." 
     -i 3 -t 18
\end{verbatim}
}

\subsubsection{Initial Implementation and Anomaly Diagnosis}

Based on the parsed literature content, the agent sets the initial weight parameters consistent with the benchmark work: $\alpha=150$ (operation assignment constraint $H_1$), $\beta=100$ (operation sequence constraint $H_2$), $\gamma=100$ (machine conflict constraint $H_3$), $\delta=15$ (makespan minimization objective $H_4$), with the complete QUBO Hamiltonian formulated as:
\begin{equation}
\min H = \alpha H_1 + \beta H_2 + \gamma H_3 + \delta H_4
\label{eq:fjsp_qubo}
\end{equation}

The definitions of each penalty term strictly follow the benchmark literature. The agent passes parameters to the modeling toolkit through a standardized interface to complete QUBO matrix construction, Ising conversion, and hardware invocation of the Coherent Ising Machine (CIM). However, the initial solution results show severe anomalies: the returned spin configuration is an all-1 solution, the optimal makespan changes directly with the setting of $T_{\text{max}}$.

Through multi-turn reasoning and comparative analysis, the central model agent autonomously locates two core errors in the initial implementation:
\begin{enumerate}
    \item \textbf{Sign Error in Ising Conversion}: The initial implementation omits the leading negative sign in the standard Ising Hamiltonian defined in Eq. (5) of this work. The correct form for CIM solving is:
    \begin{equation}
    \min E(\sigma) = -\sum_{i<j} J_{ij} \sigma_i \sigma_j - \sum_{i} h_i \sigma_i
    \label{eq:correct_ising}
    \end{equation}
    
    while the erroneous implementation uses a sign-missing form, resulting in a complete inversion of the energy landscape, turning the ground-state minimization problem into an excited-state maximization problem.
    \item \textbf{Sign Inversion of Objective Term}: The sign of the makespan minimization term $H_4$ is reversed in the initial code, changing the penalty for late completion into a 
    
    reward for extended makespan, which makes the solution completely deviate from the optimization objective.
\end{enumerate}

The central model agent autonomously performs model correction and iterative optimization under manual initial configuration and safety constraints:

\begin{enumerate}
    \item \textbf{Code Correction}: The agent guided the corrected Ising conversion code in the \texttt{nodes/ising\_conversion.py} , strictly following the conversion rules defined in Eq. (5) and Eq. (6) of this work, and fixes the sign error of the $H_4$ objective term to restore the correct optimization orientation.
    \item \textbf{Constraint Weight Iteration}: To address the persistent violations of machine conflict constraints encountered during the initial iteration stage, the agent, after conducting analysis and diagnosis, autonomously performed a gradient-based weight-tuning procedure. It gradually increased the penalty coefficient $\gamma$ from 100 to 500 to strengthen the constraint intensity, while fixing $\delta=15$ to ensure that the objective of minimizing the maximum makespan remained dominant.
    \item \textbf{Confidence Validation and Memory Mechanism}: The agent introduces a historical iteration memory module, which records the tried weight combinations in \texttt{weight\_history} to avoid repeated trials, tracks the optimal solution via \texttt{best\_makespan} and \texttt{best\_weights}, and sets \texttt{max\_history} to control memory usage, ensuring the efficiency and stability of the iterative process. Memory content derived from problem vector caches shall be excluded, so as to prevent repeated direct invocation of unconverged solutions obtained in incomplete iterations.
\end{enumerate}

\subsubsection{Validation Results and Literature Reproduction}

Based on the standardized encapsulation of the central model API, this work fully integrates a full-cycle iterative decision-making workflow, including standardized prompt construction \textit{via} \texttt{\_build\_iteration\_prompt} and standardized response parsing \textit{via} \texttt{\_parse\_iteration\_response}. The central model is invoked in each iteration round, generating detailed decision rationales, confidence evaluations (high/medium/low), and standardized JSON-formatted decision outputs, which significantly improves the standardization, intelligence, interpretability, and robustness of the decision-making process.

\begin{figure}[t!]  
\centering
\includegraphics[width=1.0\linewidth]{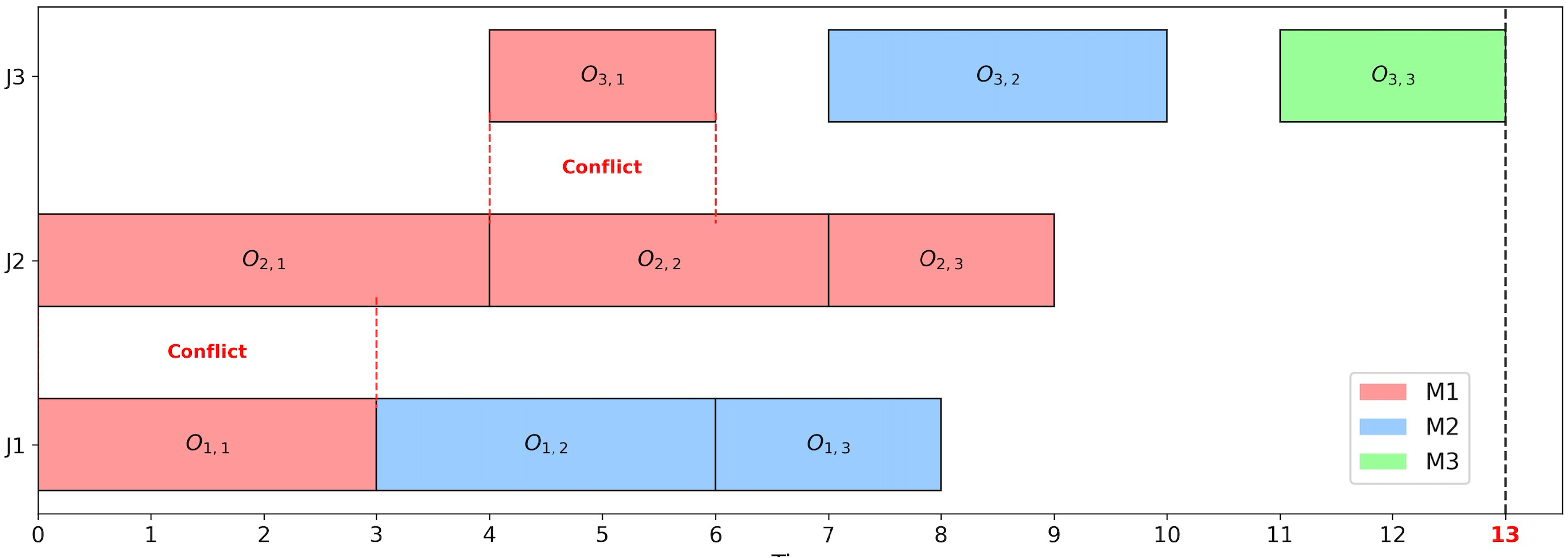}  
\caption{Machine Scheduling Conflicts }
\label{fig5}
\end{figure}

During multiple consecutive rounds of agent-guided autonomous iterations, the physical CIM hardware was successfully invoked in each run, with a single execution time of approximately 61–121 seconds and 10 feasible solutions returned per iteration. The QUBO modeling procedure was correctly implemented: the penalty coefficient  $\gamma$  was accurately transmitted, variable pruning reduces the problem scale from 513 to 264 variables, and all constraints are properly encoded.In the early stage of iteration, the coefficient  $\gamma$ is gradually tuned from 100 to 220 and then to 242. Although the CIM solver successfully yields solutions with a stable makespan of 13.0 and corresponding energies of 177107.5, 223588.75, and 180232.525 respectively, severe machine conflicts remained in the scheduling plan, including conflicting operations on Machine 1 (Fig. 5).Through diagnostic analysis, the agent further increased the penalty coefficient  $\gamma$  to 500. The optimized solution finally obtained 14.0 with an energy value of 4097000.0, and the machine conflicts had been solved (Gantt chart shown in Fig. 6a below).

\begin{figure*}[t!]  
\centering
\includegraphics[width=0.95\textwidth]{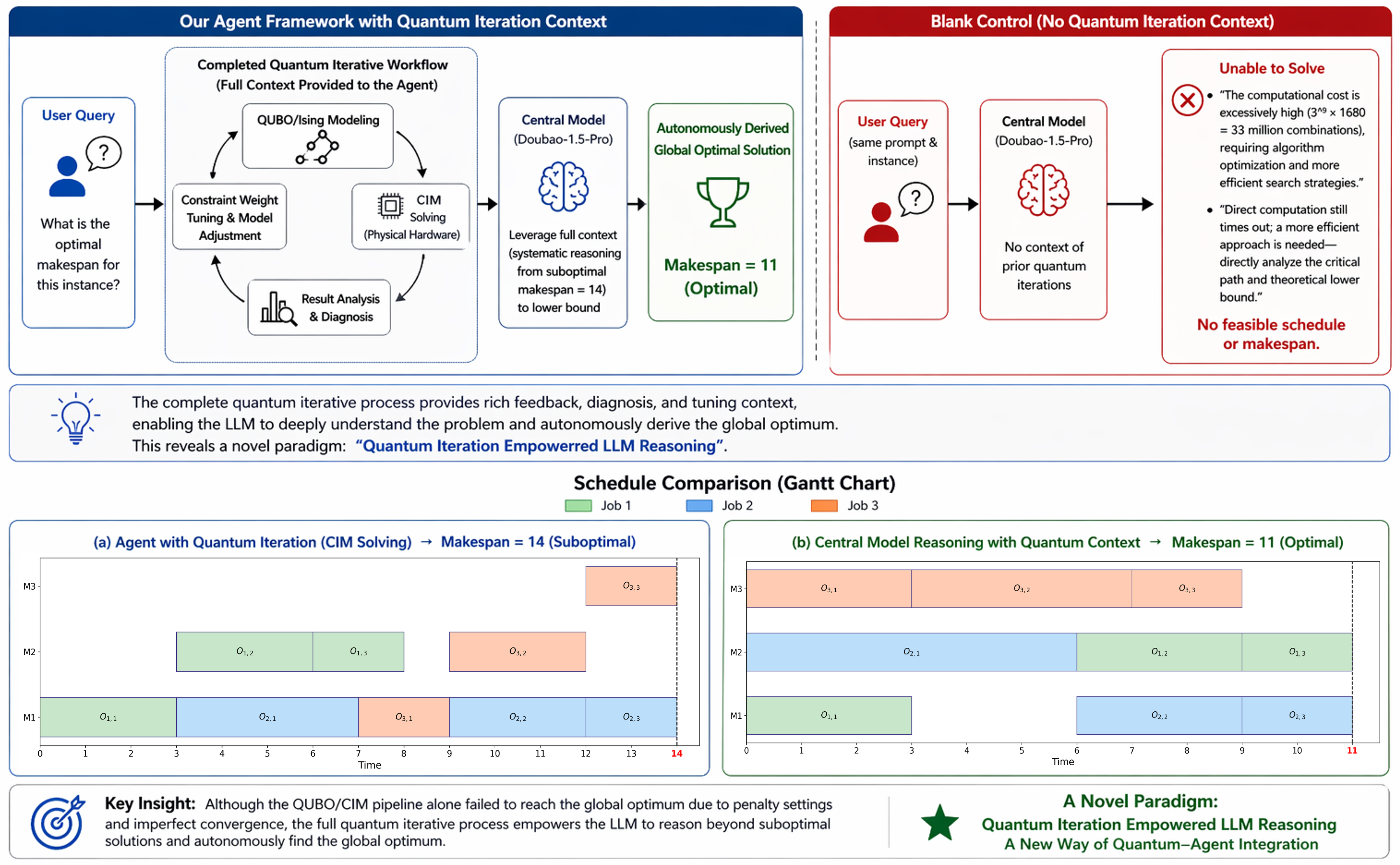}  
\caption{Quantum Iteration Empowered LLM Reasoning }
\label{fig6}
\end{figure*}

Thus, this study has successfully achieved the reproducible solution of the classic quantum optimization framework for the flexible job shop scheduling problem (FJSP) based on the physical coherent ising machine (CIM) hardware. This work verifies the capabilities of the proposed agent system in performing error diagnosis, conflict analysis, and closed-loop weight iteration in complex quantum modeling scenarios. The experimental results further demonstrate that domestic large language models exhibit strong practicability and reliability in supporting quantum computing decision-making and parameter optimization tasks.

\subsection{Unexpected Turnaround and A Novel Quantum-Enhanced Agent Paradigm}

Surprisingly, after completing the quantum optimization iteration for the aforementioned Flexible Job-Shop Scheduling Problem (FJSP), we conducted a follow-up interaction with the agent. By simply asking ``What is the optimal makespan for this instance?'', the central model leveraged the contextual information accumulated throughout the prior iterative workflow, performed systematic reasoning toward a lower makespan starting from the obtained suboptimal solution of 14, and finally autonomously derived the global optimal solution for this instance with a makespan of 11 (see the Gantt chart of the optimized scheduling scheme in Fig. 6b). Admittedly, the black‑box nature is inherent to large language models \cite{52}—here the reasoning acted more as a heuristic enumeration of known suboptimal solutions than a rigorous optimization—and we include the first‑round raw reasoning log in the Appendix (repeated rounds seemed to be primarily for validating the optimized makespan).

This experimental outcome stands in striking contrast to the output of the control groups. The control groups were set without the contextual information of the complete pre-quantum optimization iteration; instead, some underwent only basic literature-guided reasoning. When the identical central model were fed with exactly the same prompt and instance details as the experimental group, the central model in the control group consistently failed to solve the problem, only producing unhelpful responses such as ‘‘The computational cost is excessively high ($3^9 \times 1680 = 33$ million combinations), requiring algorithm optimization and more efficient search strategies'' and ‘‘Direct computation still times out; a more efficient approach is needed—directly analyze the critical path and theoretical lower bound''. It was never able to provide a feasible optimal scheduling plan or the corresponding makespan result.

Although the above results reveal that our previous case could not directly obtain the global optimal solution for this instance solely through the QUBO modeling pipeline and physical Coherent Ising Machine (CIM) hardware, limited by unreasonable penalty weight settings and incomplete convergence of the iteration process, this phenomenon conversely validates that the complete quantum optimization iterative workflow can supply effective task feedback and in-depth contextual information to the agent’s central model. More importantly, this result indicates a brand-new potential of quantum computing for supporting the reasoning of large language model (LLM)-based central models, which is entirely distinct from the conventional quantum–agent integration paradigm.

In our pre-designed agent framework, quantum computing acts merely as a callable module in the agent’s toolchain. Its application relies rigidly on precise QUBO/Ising modeling, correct Hamiltonian transformation, hardware-matched parameter tuning, and accurate solution execution; errors in any step would lead to failed solving as observed in the two cases of this study. In contrast, the newly discovered potential eliminates the need for such strict quantum computing modeling and solving procedures. Quantum computing is no longer just an isolated operational node in the agent’s execution flow. Instead, during the agent’s multiple rounds of autonomous iteration, the internal prompts and task context generated by the full quantum computing pipeline—including solving, diagnosis, and tuning—guide the agent to form an in-depth understanding of complex combinatorial optimization problems, ultimately enabling direct and autonomous problem-solving with stronger robustness. This phenomenon not only breaks through the inherent paradigm of traditional quantum computing applications but also heralds a novel agent development paradigm. We term this phenomenon quantum-iteration-empowered LLM reasoning, which opens a new paradigm where the boundary between quantum-assisted exploration and classical AI reasoning becomes mutually reinforcing.

\section{Conclusions}
This work has successfully demonstrated a practical empowerment framework for quantum Coherent Ising Machines (CIM) driven entirely by a domestic‑core agentic large model system. By integrating the Doubao‑1.5‑Pro central model with LangGraph‑orchestrated ReAct workflows, LangChain‑encapsulated quantum toolchains, and the domestically developed QBoson femtosecond‑laser‑pumped CIM hardware, we have realized a fully self‑reliant closed loop spanning natural‑language requirement parsing, autonomous QUBO/Ising model calibration, iterative constraint‑weight tuning, and physical quantum solution retrieval. The experimental validation on both a mass‑spectrometry‑inspired combinatorial task and a canonical flexible job‑shop scheduling problem confirms that domestic large models are already competent in assisting complex quantum‑computing decision‑making, including error diagnosis, conflict analysis, and literature‑guided model reproduction.

Nevertheless, the implementation journey also compels us to confront candidly the present limitations of both frontier technologies. First, despite the agent's high‑level reasoning capabilities, the underlying QUBO/Ising modeling core still relies heavily on hard‑coded implementations. The agent's principal contributions lie in understanding problem contexts, iteratively adjusting penalty weights, and selecting appropriate model structural branches—actions that are transmitted to pre‑defined, encapsulated tools via standardized parameter‑passing mechanisms. This hybrid architecture was a pragmatic necessity: fully autonomous code generation by the central model for such mathematically delicate tasks proved both token‑intensive and error‑prone under current model capabilities. Second, the physical CIM hardware, while powerful in principle, introduces its own set of fidelity barriers when deployed in fine‑grained numerical optimization scenarios. The 8‑bit integer quantization of the QUBO matrix, inherent quantum noise from optical pulse dynamics, and analog circuit imperfections collectively erode the precision of solutions that are readily achievable in numerical simulation. In our mass‑composition case, these factors manifested as persistent constraint violations and substantial mass deviations, starkly highlighting that the current hardware generation is not yet a drop‑in replacement for classical exact solvers in strict‑constraint regimes.

The latency incurred during iterative physical‑hardware invocation constitutes the very rationale for deploying an agentic architecture. The agent's role is precisely to minimize wasteful serial trial‑and‑error cycles by intelligently guiding the search in weight space, thereby amortizing the unavoidable delay over fewer, more informed queries. Yet, improving ultimate solution accuracy will demand advances on both the algorithmic and physical fronts: more sophisticated penalty‑landscape engineering, higher‑precision analog coupling, and crucially, the scaling up of qubit counts and quantization bit‑widths to faithfully capture the dynamic range required by real‑world applications.

Encouragingly, the most profound insight of this study emerges from an unexpected turnaround. When the quantum‑assisted iterative workflow did not directly converge to the global optimum under the initial weight‑tuning and hardware‑precision constraints, the very process of quantum‑informed trial, diagnosis, and contextual feedback endowed the central model with a deep, structured comprehension of the problem. This accumulated "quantum‑iteration experience" subsequently enabled the model, in a follow‑up dialogue devoid of further hardware calls, to autonomously reason its way to a provably superior solution that the identical model, absent this prior quantum‑contextual exposure, was entirely incapable of producing. This phenomenon reveals a paradigm that transcends the initially envisioned quantum‑as‑a‑tool model: quantum computing iteration can reciprocally empower large language model reasoning. The agent does not merely call quantum hardware; it learns from the hardware's feedback loops, internalizing domain‑specific heuristics that fortify its native problem‑solving faculties. Such a symbiosis, wherein incomplete or even suboptimal quantum runs still catalyze robust cognitive enhancement in the agent, hints at a new developmental trajectory where the boundary between classical AI reasoning and quantum computational exploration becomes increasingly porous and mutually reinforcing.

In summary, this work establishes a solid technological foundation for all‑domestic ‘‘Large Model + Quantum CIM" integration, while transparently documenting the current gaps in autonomous code generation and analog‑hardware precision. More importantly, it unveils a promising, previously unanticipated avenue in which the iterative struggle with quantum constraints serves not as a dead end but as a unique training regimen for the large‑model mind. As both domestic quantum hardware and foundation models continue their rapid evolution, this reciprocal empowerment loop may well define a resilient and continually self‑improving paradigm for next‑generation intelligent computing.

\appendix

\subsection*{Appendix }

\subsubsection*{Iterative Optimization of QUBO Modeling in Section 3.1}

We initially constructed a QUBO model based on binary count encoding, where each amino acid was represented by a 5-bit binary variable to characterize its quantity, defined as $x_{a,k} \in \{0,1\}$ representing the $k$-th binary bit of the copy number of the $a$-th amino acid, the total copy number of the $a$-th amino acid was $N_a = \sum_{k=0}^{4} 2^k x_{a,k}$, and the total QUBO energy function was constructed by combining the total mass matching constraint and the sequence length constraint, which was expressed as:

{\small
\[
H = A\left(\sum_{a=1}^{20} m_a N_a - M_{\text{target}}\right)^2 + E\left(\sum_{a=1}^{20} N_a - L_{\text{mid}}\right)^2
\]
}

where $A$ and $E$ were the weight coefficients of the mass constraint and the length constraint respectively, $L_{\text{mid}}$ was the midpoint of the target sequence length range, after expanding the energy function, the linear term coefficient is $A \cdot 2m_a 2^k (m_a 2^k - M_{\text{target}}) + E \cdot 2 \cdot 2^k (2^k - L_{\text{mid}})$ and the quadratic term coefficient was $2A m_a m_{a'} 2^k 2^{k'} + 2E \cdot 2^k 2^{k'}$, the agent carried out coefficient normalization processing on the model to adapt to the hardware solving range of the coherent Ising machine.

After actual solving by CIM and result analysis, we found that the sequence length constraint failed to take effect effectively, and further reasoning by agent and analysis showed that the coefficient scaling process led to the overall weight of the constraint term being too low, and the encoding method that multiplies the binary count weight by the amino acid mass greatly suppressed the numerical value of most linear and quadratic terms, making most coefficients approach zero, resulting in the solver only being able to generate solutions composed of amino acids with a large number of copies and a large molecular weight.

To evidence this problem, we tried to use a masking mechanism to increase the weight of target amino acids and strengthen the contribution of key terms in the energy function, the results confirmed coefficient suppression caused by the coupling of count encoding and mass parameters.

Based on the above analysis and iterative verification, we finally made a decisive decision to abandon the binary count encoding and switch to the position-based one-hot encoding modeling method, redefined the binary variable $x_{s,a} \in \{0,1\}$ which means that the $s$-th position of the sequence selects the $a$-th amino acid, reconstructed the QUBO model with the position one-hot constraint and the total mass matching constraint as the core, the position one-hot constraint ensured that each position has and only one amino acid selected, and its penalty term is:

\[
\lambda_{\text{pos}} \sum_{s=1}^{S} \left(1 - \sum_{a} x_{s,a}\right)^2
\]

where $\lambda_{\text{pos}}$ was the penalty weight of the position constraint, the total mass matching constraint minimized the deviation between the total mass of the sequence and the target mass, and its penalty term was:

\[
\lambda_{\text{mass}} \left(\sum_{s=1}^{S} \sum_{a} m_a x_{s,a} - M_{\text{target}}\right)^2
\]
where $\lambda_{\text{mass}}$ was the penalty weight of the mass constraint, the total QUBO model was the sum of the two penalty terms, namely:
{\small
\[
H = \lambda_{\text{pos}} \sum_{s=1}^{S} \left(1 - \sum_{a} x_{s,a}\right)^2 + \lambda_{\text{mass}} \left(\sum_{s=1}^{S} \sum_{a} m_a x_{s,a} - M_{\text{target}}\right)^2
\]
}

This modeling method eliminated the coupling between count-based encoding and mass parameters, avoided the coefficient suppression issue induced by scaling, and retained clear physical interpretability. The agent assisted the construction of the optimized QUBO matrix, the invocation of quantum hardware, the analysis of all CIM solution outputs, and the iterative adjustment of constraint weights, and finally obtained a stable and reasonable solution.

\subsubsection*{Detailed Formulation of the Final QUBO Model in Section 3.1}

The final position-based one-hot encoding scheme yields a well-defined QUBO formulation that balances both the one-hot constraint and mass matching requirement, eliminating the coefficient suppression issue of the previous count-based encoding. The complete QUBO objective function is the weighted sum of two penalty terms, as defined below:

\[
\min_{\mathbf{x}} \quad H_{\mathrm{QUBO}} = H_{\mathrm{pos}} + H_{\mathrm{mass}}
\]

\paragraph{1. Position One-Hot Constraint ($H_{\mathrm{pos}}$)}

This term enforces that exactly one amino acid is selected at each sequence position. The penalty function is defined as:

\[
H_{\mathrm{pos}} = \lambda_{\mathrm{pos}} \sum_{s=1}^{S} \left(1 - \sum_{a \in \mathcal{A}} x_{s,a}\right)^2
\]

where $S$ is the total number of sequence positions, $\mathcal{A}$ denotes the set of candidate amino acids, and $x_{s,a} \in \{0,1\}$ is the binary variable indicating whether amino acid $a$ is selected at position $s$. $\lambda_{\mathrm{pos}}$ is the penalty weight for the one-hot constraint.

Expanding the quadratic form and simplifying using $x_{s,a}^2 = x_{s,a}$ (valid for binary variables), the term becomes:

\[
H_{\mathrm{pos}} = \lambda_{\mathrm{pos}} \sum_{s=1}^{S} \left[ -\sum_{a \in \mathcal{A}} x_{s,a} + 2 \sum_{\substack{a_1 < a_2 \\ a_1,a_2 \in \mathcal{A}}} x_{s,a_1} x_{s,a_2} + 1 \right]
\]

The constant term (independent of $\mathbf{x}$) can be omitted in optimization, as it does not affect the optimal solution.

\paragraph{2. Total Mass Matching Constraint ($H_{\mathrm{mass}}$)}

This term minimizes the deviation between the total mass of the selected sequence and the target mass $M_{\mathrm{target}}$. The penalty function is defined as:

\[
H_{\mathrm{mass}} = \lambda_{\mathrm{mass}} \left( \sum_{s=1}^{S} \sum_{a \in \mathcal{A}} m_a x_{s,a} - M_{\mathrm{target}} \right)^2
\]

where $m_a$ is the monoisotopic mass of amino acid $a$, and $\lambda_{\mathrm{mass}}$ is the penalty weight for the mass matching constraint.

\paragraph{3. QUBO Standard Form and Matrix Construction}

The combined objective function is reformulated into the standard QUBO form:

\[
\min_{\mathbf{x} \in \{0,1\}^N} \quad \mathbf{x}^T Q \mathbf{x}
\]

where $N = S \times |\mathcal{A}|$ is the total number of binary variables. The entries of the QUBO matrix $Q$ are derived by combining the linear and quadratic coefficients from both penalty terms:

\begin{itemize}

    \item \textbf{Linear terms (diagonal elements $Q_{ii}$):}
    
    For a variable $x_{s,a}$ (indexed by $(s,a)$), the diagonal coefficient is:
    
    \[
    Q_{(s,a),(s,a)} = \lambda_{\mathrm{pos}} \cdot (-1) + \lambda_{\mathrm{mass}} \cdot \left( m_a^2 - 2 M_{\mathrm{target}} m_a \right)
    \]
    
    \item \textbf{Quadratic terms (off-diagonal elements $Q_{i,j}, i < j$):}
    \begin{itemize}
        \item For two variables at the same position ($s$) but different amino acids ($a_1 \neq a_2$):
        \[
        Q_{(s,a_1),(s,a_2)} = \lambda_{\mathrm{pos}} \cdot 2
        \]
        \item For any two distinct variables (across different positions or different amino acids):
        \[
        Q_{(s_1,a_1),(s_2,a_2)} = \lambda_{\mathrm{mass}} \cdot 2 m_{a_1} m_{a_2}
        \]
    \end{itemize}
\end{itemize}

This explicit matrix formulation is directly used to construct the QUBO instance after coefficient normalization process for solving on the CIM hardware.

\subsubsection*{Details of Flexible Job-Shop Scheduling Problem (FJSP) in Section 3.2}
This section presents the formal definition, mathematical modeling, QUBO formulation, Ising model conversion, and variable pruning method for FJSP, strictly following the framework proposed by Fu et al.  \cite{24}.

\paragraph*{1. Mixed-Integer Programming (MIP) Model}
We first introduce the classical MIP model for FJSP, which serves as the basis for QUBO formulation. The decision variables are defined as:
\begin{itemize}
    \item $x_{i,j,h} \in \{0,1\}$: Equals 1 if operation $o_{j,h}$ is processed on machine $i$, 0 otherwise.
    \item $y_{i,j,h,j',h'} \in \{0,1\}$: Equals 1 if operation $o_{j,h}$ is processed before $o_{j',h'}$ on machine $i$, 0 otherwise.
\end{itemize}

The MIP model is formulated as:
\begin{equation}
\sum_{i \in M_{j,h}} x_{i,j,h} = 1, \quad \forall j \in J, h = 1,...,h_j
\label{eq:mip1}
\end{equation}

\begin{equation}
\begin{aligned}
& s_{j,h} + x_{i,j,h} \cdot p_{i,j,h} \leq c_{j,h}, \\
& \forall i \in M_{j,h},\ j \in J,\ h = 1,\dots,h_j
\end{aligned}
\label{eq:mip2}
\end{equation}

\begin{equation}
c_{j,h} \leq s_{j,h+1}, \quad \forall j \in J, h = 1,...,h_j-1
\label{eq:mip3}
\end{equation}

\begin{equation}
\begin{aligned}
s_{j,h} + p_{i,j,h} \leq s_{j',h'} + L\left(1 - y_{i,j,h,j',h'}\right),\\
\quad \forall i \in M_{j,h} \cap M_{j',h'}, j \neq j'
\end{aligned}
\label{eq:mip4}
\end{equation}

\begin{equation}
c_{j,h_j} \leq C_{\text{max}}, \quad \forall j \in J
\label{eq:mip5}
\end{equation}

\begin{equation}
s_{j,h}, c_{j,h} \geq 0, \quad \forall j \in J, h = 1,...,h_j
\label{eq:mip6}
\end{equation}

\begin{equation}
x_{i,j,h}, y_{i,j,h,j',h'} \in \{0,1\}, \quad \forall i,j,h,j',h'
\label{eq:mip7}
\end{equation}

\noindent 

where:
- Constraint \eqref{eq:mip1}: Each operation is assigned to exactly one eligible machine.
- Constraints \eqref{eq:mip2}-\eqref{eq:mip3}: Ensure non-overlapping processing of sequential operations within the same job.
- Constraint \eqref{eq:mip4}: Prevents temporal conflicts of operations assigned to the same machine.
- Constraint \eqref{eq:mip5}: Defines the makespan $C_{\text{max}}$ as the maximum completion time of all jobs.

\paragraph*{2. Quadratic Unconstrained Binary Optimization (QUBO) Model}

To adapt the FJSP for quantum computing via a coherent Ising machine (CIM), we transform the MIP model into a QUBO formulation by discretizing the timeline and encoding all constraints as penalty terms in the objective Hamiltonian. The core binary variable for QUBO modeling is defined as:

\begin{equation}
\scalebox{0.85}{$\displaystyle
k_{i,t,o_{j,h}} = 
\begin{cases}
1, & o_{j,h} \text{ starts on machine } i \text{ at time } t \in T \\
    0, & \text{otherwise}
\end{cases}
 $}
\label{eq:qubo_var}
\end{equation}

The complete QUBO model is formulated as a Hamiltonian minimization problem, where hard constraints are encoded as quadratic penalty terms with predefined weights, and the makespan minimization is the linear objective term:

\begin{equation}
\min H = \alpha H_1 + \beta H_2 + \gamma H_3 + \delta H_4
\label{eq:fjsp_qubo_hamiltonian}
\end{equation}

where $\alpha, \beta, \gamma, \delta$ are positive penalty coefficients for constraints and the objective term, respectively. The canonical values from the benchmark work are $\alpha=150$, $\beta=100$, $\gamma=100$, $\delta=15$.

Each term in the Hamiltonian is defined as follows:
1. Operation Assignment Constraint ($H_1$): Ensures each operation is assigned to exactly one machine at exactly one start time.
\begin{equation}
H_1 = \sum_{o_{j,h} \in O_j} \left(1 - \sum_{t \in T} \sum_{i \in M_{j,h}} k_{i,t,o_{j,h}}\right)^2
\label{eq:h1}
\end{equation}

2. Operation Sequence Constraint ($H_2$): Guarantees non-overlapping sequential operations within the same job, i.e., an operation can only start after the previous operation of the same job is completed.

\begin{equation}
\scalebox{0.85}{$\displaystyle
H_2 = \sum_{j \in J} \sum_{h=1}^{h_j-1} \sum_{t \in T} \sum_{i \in M_{j,h}} k_{i,t,o_{j,h}} \cdot \sum_{t' < t + p_{i,j,h}} \sum_{i' \in M_{j,h+1}} k_{i',t',o_{j,h+1}}
 $}
\label{eq:h2}
\end{equation}

3. Machine Conflict Constraint ($H_3$): Prevents multiple operations from being processed on the same machine at overlapping time intervals.

\begin{equation}
\scalebox{0.88}{$\displaystyle
H_3 = \sum_{(j,h,j',h',t,t') \in G \cup H} \sum_{i \in M_{j,h} \cap M_{j',h'}} k_{i,t,o_{j,h}} k_{i,t',o_{j',h'}}
$}
\label{eq:h3}
\end{equation}

where the sets are defined as:

\begin{align*}
\footnotesize
G &= \left\{(j,h,j',h',t,t') \mid j,j' \in J,\ j \neq j',\ t,t' \in T,\right. \\
  &\qquad \left. 0 \leq t - t' \leq p_{i,j,h}\right\}, \\
H &= \left\{(j,h,j',h',t,t') \mid j,j' \in J,\ j \neq j',\ t,t' \in T,\right. \\
  &\qquad \left. 0 \leq t' - t \leq p_{i,j',h'}\right\}.
\end{align*}

4. Makespan Minimization Objective ($H_4$): Penalizes late completion of operations to minimize the overall makespan.
\begin{equation}
H_4 = \sum_{j \in J} \sum_{i \in M_{j,h}} \sum_{t \in T} k_{i,t,o_{j,h}} \cdot \left(t + p_{i,j,h} - P_{o_{j,h}}\right)
\label{eq:h4}
\end{equation}

where $P_{o_{j,h}}$ is the minimum predecessor time of operation $o_{j,h}$, defined as the sum of the minimum processing times of all preceding operations of $o_{j,h}$:
\begin{equation}
P_{o_{j,h}} = \sum_{l_{o_{j,h'}} < l_{o_{j,h}}} \min_{i' \in M_{j',h'}} p_{i',j',h'}
\label{eq:predecessor_time}
\end{equation}

and $l_{o_{j,h}}$ denotes the sequence order of operation $o_{j,h}$ within its job.

\paragraph*{3. Conversion from QUBO to Ising Model}
The CIM solver requires the problem to be formulated as an Ising model with spin variables $\sigma_i \in \{-1, +1\}$. The standard QUBO form is given in Eq.~\eqref{eq:qubo_full_form}:
\begin{equation*}
\min_{\boldsymbol{x} \in \{0,1\}^n} \boldsymbol{x}^T Q \boldsymbol{x} + \boldsymbol{c}^T \boldsymbol{x}
\end{equation*}

where $Q \in \mathbb{R}^{n \times n}$ is the quadratic coefficient matrix, and $\boldsymbol{c} \in \mathbb{R}^n$ is the linear coefficient vector.

We perform the variable substitution to map binary QUBO variables to Ising spin variables:
\begin{equation}
\sigma_i = 2x_i - 1 \quad \Leftrightarrow \quad x_i = \frac{\sigma_i + 1}{2}
\label{eq:qubo_ising_mapping}
\end{equation}

where $\sigma_i \in \{-1, +1\}$ is the spin variable for the $i$-th qubit in the CIM.

Substituting \eqref{eq:qubo_ising_mapping} into \eqref{eq:qubo_full_form}, we derive the standard Ising Hamiltonian for minimization:
\begin{equation}
\min E(\boldsymbol{\sigma}) = -\sum_{i<j} J_{ij} \sigma_i \sigma_j - \sum_{i} h_i \sigma_i
\label{eq:standard_ising}
\end{equation}

where:
- $J_{ij}$ is the coupling coefficient between spin $i$ and spin $j$, derived from the quadratic terms of the QUBO matrix $Q$;
- $h_i$ is the external magnetic field coefficient for spin $i$, derived from the linear terms of the QUBO model.

The Ising matrix constructed from the QUBO model is directly fed into the CIM solver for ground-state computation, where the minimum energy state corresponds to the optimal FJSP scheduling solution.

\paragraph*{4. Variable Pruning Approach}
To reduce the number of qubits required for quantum computation, we implement two variable pruning strategies to eliminate invalid binary variables $k_{i,t,o_{j,h}}$:

1. Machine Eligibility Pruning: For any operation $o_{j,h}$ and machine $i$ that is not in the eligible machine set $M_{j,h}$, the variable is fixed to 0 for all time steps:
\begin{equation}
k_{i,t,o_{j,h}} = 0, \quad \forall t \in T, i \notin M_{j,h}, j \in J
\label{eq:prune1}
\end{equation}

2. Time Window Pruning: An operation can only start processing after all its preceding operations are completed, and must finish before the minimum start time of its subsequent operations. We fix the variable to 0 for all time steps outside the valid time window:
\begin{equation}
k_{i,t,o_{j,h}} = 0, \quad \forall t \in T \text{ with } t < P_{o_{j,h}} \text{ or } t > S_{o_{j,h}}
\label{eq:prune2}
\end{equation}

where $P_{o_{j,h}}$ is the minimum predecessor time defined in \eqref{eq:predecessor_time}, and $S_{o_{j,h}}$ is the maximum allowable start time, defined as the sum of the minimum processing times of all successor operations of $o_{j,h}$:
\begin{equation}
S_{o_{j,h}} = T_{\text{max}} - \sum_{l_{o_{j,h'}} > l_{o_{j,h}}} \min_{i' \in M_{j',h'}} p_{i',j',h'}
\label{eq:successor_time}
\end{equation}

This pruning approach significantly reduces the number of qubits required. For the 3×3 FJSP instance, the number of qubits is reduced from 513 to 264 after pruning, greatly improving the computational efficiency of the CIM solver.

\subsubsection*{LLM’s Reasoning in Section 3.3}
While the system logs iterative optimization data (\textit{e.g.}, \texttt{weight\_history}, \texttt{best\_makespan}) in memory, the final serendipitous reasoning was only captured \textit{via} console output and not pre-logged. We will reproduce and upload these records to GitHub after repeating the experiment if necessary. The full first-round reasoning dialogue follows below.

\vspace{3pt}
\noindent\rule{\linewidth}{0.5pt}
\vspace{2pt}
{
\footnotesize
-To solve the 3$\times$3 Flexible Job-Shop Scheduling Problem (FJSP), we first clarify the problem definition: it involves 3 jobs, each containing 3 operations, where each operation can be processed on a selected machine from M1, M2, and M3. The optimization objective is to minimize the makespan (total completion time).

-We first organize the processing times clearly: for example, the processing times of each operation of Job 1 on M1, M2, and M3 are 3, 4, and 5, respectively. The scheduling rules for the flexible job shop are also specified: operations of each job must follow a sequential processing order; operations on the same machine cannot overlap; the objective is to minimize the makespan; and machines with shorter processing times are prioritized to avoid machine conflicts and satisfy operation precedence constraints.

-We start by analyzing the shortest processing times for each operation of Job 1, select machines accordingly, and derive its completion time as 8, along with the corresponding machine utilization profile. Next, we analyze Job 2, identify the shortest processing time for each operation, account for conflicts with operations of Job 1, and deduce its completion time and machine usage. We list the available operations and their corresponding processing times on M2 and M3, calculating the total processing times as 37 and 42, respectively. We then attempt to assign machines to each operation, schedule their start and end times, and compute the makespan. For instance, we first propose partial scheduling schemes for Job 1 and Job 2.

-We assign Operation $O_{2,3}$ to M1, calculating its start time as 8 and completion time as 10, yielding a completion time of 10 for Job 2. For Job 3, we sequentially derive the start and end times of each operation on different machines: $O_{3,1}$ on M1 starts at 3 and finishes at 5; $O_{3,2}$ on M2 starts at 8 and finishes at 11; $O_{3,3}$ on M3 starts at 11 and finishes at 13, resulting in a completion time of 13 for Job 3. We sort the operations on each machine along the time axis and find no overlaps on M1, M2, or M3, with all operation precedence constraints satisfied. The resulting makespan is 13, better than the user's log. However, since the user asks for the globally optimal makespan via exhaustive analysis, we attempt to find a smaller makespan (e.g., 12 or 11) to explore the optimal solution.

-We first test the feasibility of a makespan of 12. We calculate the total processing time of all operations by selecting the shortest processing time for each job: 8 for Job 1, 9 for Job 2, and 7 for Job 3, giving a total processing time of 24 and an average machine load of 8 across the three machines. The theoretical lower bound must account for operation precedence and machine conflicts. Recalculating the total processing time of each job in sequence confirms 8 for Job 1, 9 for Job 2, and 7 for Job 3, meaning the makespan is at least 9. In practice, however, machine conflicts lead to a longer completion time. Given an average machine processing time of 8, the theoretical lower bound is $\max(9, 8) = 9$, but practical constraints from operation sequences and machine conflicts raise the actual lower bound. We then proceed to search for a schedule with a makespan of 12.

-Using a conventional assignment strategy, we first schedule the operations of Job 2: $O_{2,1}$ on M1 from 3 to 7, and $O_{2,2}$ on M3 from 7 to 12. Since $O_{2,3}$ must be processed after $O_{2,2}$ and both M1 and M3 are occupied, this scheme results in Job 2 finishing after 12 and is thus infeasible. We therefore revise the operation assignment: $O_{2,1}$ on M3 from 0 to 5, $O_{2,2}$ on M2 from 6 to 10, and $O_{2,3}$ on M1 from 10 to 12, yielding a completion time of 12 for Job 2.

-We re-examine all operations and first derive the processing times for Job 1: $O_{1,1}$ on M1 from 0 to 3, $O_{1,2}$ on M2 from 3 to 6, and $O_{1,3}$ on M2 from 6 to 8, finishing at 8. For Job 2, the original assignment of $O_{2,1}$ on M1 from 3 to 7 and $O_{2,2}$ on M3 from 7 to 12 is infeasible because $O_{2,3}$ must follow $O_{2,2}$ and M3 remains occupied. We reassign operations: $O_{2,1}$ on M2 from 0 to 6, $O_{2,2}$ on M1 from 6 to 9, and $O_{2,3}$ on M1 from 9 to 11, so Job 2 finishes at 11. We then adjust $O_{1,2}$ of Job 1 to start at 6 and finish at 9 to resolve the conflict with $O_{2,1}$.

-We assign $O_{1,3}$ to M2, starting at 9 and finishing at 11, resulting in a completion time of 11 for Job 1. We re-derive the operations of Job 3: $O_{3,1}$ on M3 from 0 to 3, $O_{3,2}$ on M3 from 3 to 7, and $O_{3,3}$ on M3 from 7 to 9, so Job 3 finishes at 9. Sorting the operations on each machine reveals no conflicts and full compliance with operation precedence constraints. The makespan is calculated as $\max(11, 11, 9) = 11$. We then verify the rationality of this result.

-We sequentially check the operation sequences and processing times of all jobs, confirming that Job 1, Job 2, and Job 3 all satisfy precedence constraints and that the processing times are correct. The total processing times on each machine are 8 for M1, 11 for M2, and 9 for M3, with completion times of 11 on M1, 11 on M2, and 9 on M3. The makespan is 11, which is shorter than the previous 13. We then verify the feasibility of this schedule. We doubt the validity of the derived makespan of 11 and compare it with the makespan of 13 recorded in the user's log. We double-check the processing times of each operation of Job 1, Job 2, and Job 3 on different machines, confirming their correctness to lay a foundation for further exploring the optimal solution.

-We remain skeptical of the makespan of 11 and re-inspect the operation sequences and processing times of all jobs. We confirm that the processing times and sequences of all operations on different machines are correct, with no overlapping machine conflicts, revalidating the feasibility of this schedule. We first verify that no overlaps exist on M3 and that all operation sequences are correct, confirming the current makespan of 11 is indeed shorter than the previous result. We then test whether a makespan of 10 is feasible. Calculations show that Job 1 and Job 2 finish at 8 and 10, respectively, but operation conflicts delay Job 3 such that its completion time exceeds 10, rendering this scheme infeasible.

-We try alternative machine assignments for Job 3: setting $O_{3,2}$ on M3 (processing time 4) results in a start time of 5 and end time of 9, but assigning $O_{3,3}$ to M3 (processing time 2) gives a start time $\geq$ 9 and end time of 11, which exceeds 10. We reassign Job 3's operations: $O_{3,1}$ on M2 (4), $O_{3,2}$ on M1 (5), and $O_{3,3}$ on M3 (2), which still results in $O_{3,3}$ finishing after 10. Another trial schedule adjusts Job 1's operations across different machines; when $O_{1,3}$ is assigned to M2 (2), Job 1 finishes at 9, with no conflicts on M2.

-We assign Job 1's operations entirely on M2, yielding a completion time of 9. Job 2 is scheduled on M1 and M3: $O_{2,1}$ from 0 to 4, $O_{2,2}$ from 4 to 7, and $O_{2,3}$ from 7 to 10, finishing at 10. For Job 3, assigning $O_{3,1}$ to M1 (2) gives a start time of 7 and end time of 9, but assigning $O_{3,2}$ to M3 (4) or M2 (3) results in a start time $\geq$ 9 and end time exceeding 10, which is infeasible. Assigning $O_{3,1}$ to M3 (3), $O_{3,2}$ to M1 (5), and $O_{3,3}$ to M3 (2) also causes $O_{3,3}$ to finish after 10. Analysis shows that the total processing time of $O_{3,2}$ and $O_{3,3}$ for Job 3 is 6, plus 3 for $O_{3,1}$, summing to 9. Due to machine conflicts, achieving a makespan of 10 is highly impractical.

-We re-examine the earlier schedule with a makespan of 11, confirming that the processing and start times of all operations for Job 1 and Job 2 meet the required constraints. We explore further optimality and test new schedules, all of which still yield a makespan of 11. We first check that $O_{3,3}$ on M3 runs from 9 to 11, with no machine conflicts. However, a re-examination reveals overlapping processing times between $O_{1,2}$ and $O_{2,1}$ on M3. Adjusting the start time of $O_{1,2}$ to 5 pushes its completion time to 13, returning the makespan to 13. Upon further verification, we confirm that in the original schedule with a makespan of 11, the processing times of $O_{2,1}$ and $O_{1,2}$ on M2 are consecutive but non-overlapping. Consecutive processing on the same machine is permitted as long as no overlap exists and the end time of one operation matches the start time of the next.

-We assess whether the schedule with a makespan of 11 is optimal and test another configuration with the same makespan. Calculations for revised assignments of Job 1 and Job 2 both yield completion times of 10. Testing different machine options for Job 3's operations shows that assigning $O_{3,3}$ to M2 (6), M3 (2), or M1 (3) all result in completion times exceeding 11. Analysis indicates that the shortest processing time for $O_{3,3}$ is 2 on M3, but occupation of M3 delays its processing.

-We reaffirm the feasibility of the earlier schedule with a makespan of 11, verifying that all operations of Job 1, Job 2, and Job 3 satisfy sequence constraints and machine non-overlap requirements, and examine whether this represents the optimal solution. We then hypothesize a makespan of 10 and analyze the feasibility of all operations of Job 1 and Job 2 finishing by 10. Analysis suggests the start time of Job 1's operations must be $\leq$ 2 and end time $\leq$ 5, while all operations of Job 2 finish by 10, with no obvious conflicts. Further exploration is still required for a schedule with a makespan of 10.

-We first analyze the processing times of $O_{1,1}$ and $O_{2,1}$ on M1 and find they overlap, meaning they cannot both be assigned to M1. We adjust the assignment: $O_{1,1}$ on M1 from 0 to 3 and $O_{2,1}$ on M3 from 0 to 5 to avoid conflicts, calculating a completion time of 10 for Job 2. We then derive the processing times of $O_{1,2}$ and $O_{1,3}$ on M2, confirming Job 1 finishes between 8 and 10 (e.g., $O_{1,2}$ from 3 to 6 and $O_{1,3}$ from 6 to 8). Finally, we analyze Job 3's operations and find that assigning $O_{3,1}$ to M2 causes a conflict with $O_{1,2}$, concluding that $O_{3,1}$ cannot be assigned to M2.

-We test assigning $O_{3,1}$ to M1 (2) and M3 (3) separately. Assigning $O_{3,1}$ to M1 gives a start time $\geq$ 3 and end time of 5, which is feasible; assigning it to M3 results in processing from 0 to 3, also feasible. We then derive schedules for $O_{3,2}$ on M3 (4), M2 (3), and M1 (5), but all result in start times exceeding 10 due to machine occupation, making them infeasible. Another attempt assigning $O_{3,2}$ to M1 (5) and $O_{3,3}$ to M1 (3) also causes $O_{3,3}$ to finish after 10. In summary, assigning $O_{3,1}$ to M1 (2) or M3 (3), $O_{3,2}$ to M1 (5), and $O_{3,3}$ to M1 (3) yields processing times for Job 3 that remain within 10, proving this assignment feasible.

-We recalculate the completion times of Job 2 and Job 3 under the new assignment, obtaining a makespan of 11, consistent with earlier results. This confirms the existence of a feasible schedule with a makespan of 11. We speculate on the reason for the makespan of 14 in the user's log: it likely represents a suboptimal solution obtained either due to incomplete solution convergence or inappropriate penalty weight settings, with adjusted weights yielding this suboptimal result.

-We explore whether a makespan smaller than 11 (e.g., 10) is achievable and conclude that it is not. Calculating the total sequential processing time of each job --- 8 for Job 1, 9 for Job 2, and 7 for Job 3 --- establishes a theoretical lower bound of at least 9. Machine conflicts extend the practical makespan, and while 11 is feasible, 10 is unattainable. We therefore propose that 11 is likely the optimal makespan. We test another schedule aiming for a makespan of 11 but find that assigning $O_{2,3}$ of Job 2 to M3 or M1 results in completion times exceeding 11, rendering this scheme invalid. We confirm the uniqueness of the feasible 11-makespan schedule and recheck all operation times and sequences, finding no errors.

-We verify the schedule for Job 3's operations, confirming a valid sequence of completion times: 3 $\rightarrow$ 7 $\rightarrow$ 9, satisfying all constraints. This schedule is feasible with a makespan of 11, which is superior to the 14 reported in the user's log. We conclude that the user's earlier solution did not reach the global optimum, likely due to inappropriate weight settings or insufficient iterative convergence.

}
\noindent\rule{\linewidth}{0.5pt}
\vspace{3pt}

\subsection*{Acknowledgment}
We sincerely acknowledge Academician Ding Hong from Fudan University for his insightful inspirations and valuable academic guidance before this work. We are grateful to QBoson Quantum Technology Co., Ltd. for providing technical support and access to the Coherent Ising Machine (CIM) hardware platform. We also thank Liu Dinghao from China Pharmaceutical University for constructive academic discussions and generous assistance in this research. This work was financially supported by Tsinghua University through multiple research practice programs.

\subsection*{Declaration of competing interest}

The authors have no competing interests to declare that are relevant to the
content of this article.\\

\subsection*{Data Availability}
We will release and continuously update the code and all data required to verify the presented conclusions on Github.



\subsection*{Author biography}

\begin{biography}[author1]{Wang Rui} (ORCiD: 0000-0002-2362-3214) is an Assistant Research Fellow (Postdoctoral Researcher) at the Department of Chemical Engineering, Tsinghua University. He earned his Ph.D. in Chemistry from Tsinghua in 2024 and has published multiple first‑ or corresponding‑author peer-reviewes papers. His research interests center on the deep integration of AI with chemistry and life sciences, particularly in the emerging field of AI for Science.
\end{biography}

\begin{biography}[author2]{Lu Diannan}  is a Professor and Ph.D. Supervisor at the Department of Chemical Engineering, Tsinghua University. Lu earned a B.Eng. (2000) and a Ph.D. (2006) in Chemical Engineering from Tsinghua University and currently also serves at the Graduate School of Tsinghua University. Recognitions include the National Excellent Doctoral Dissertation Award (2008), the Basic Research Talent Program of Tsinghua University (2012), and the Beijing Young Talent Program (2013). Professional society memberships span the Chinese Society for Environmental Sciences (2017–), the Chinese Chemical Society (2016–), the Metallurgical Physical Chemistry Committee of the Chinese Society for Metals (2015–), the American Institute of Chemical Engineers (2006–), and the Chemical Industry and Engineering Society of China (2006–). Research interests center on thermodynamics of complex systems and molecular simulation, with a focus on functional nanochannels, biocatalysts, macromolecular drugs, and soil bioremediation. Lu has made prominent contributions to chemical engineering thermodynamics education and AI‑empowered higher education, including teaching national‑level elite courses and developing the first AI teaching assistant system.

\end{biography}

\vspace*{2.6em}

\subsection*{Graphical abstract}

\vspace{1em}
\begin{center}
\includegraphics[width=1.0\linewidth]{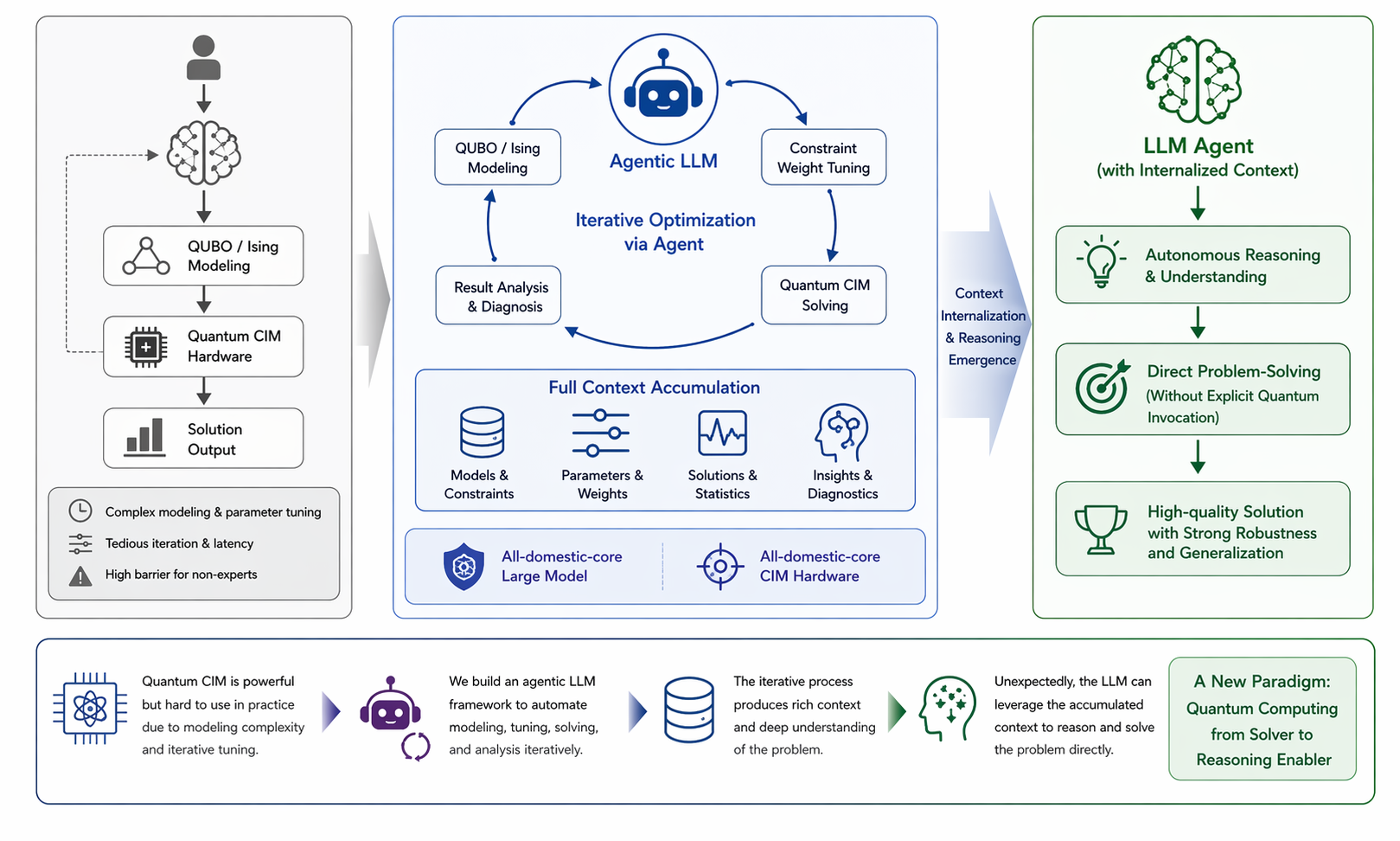}
\end{center}

\end{document}